\documentclass[lettersize,journal]{IEEEtran}
\usepackage{amsmath,amsfonts}
\usepackage[linesnumbered, ruled]{algorithm2e}
\usepackage{array}
\usepackage[caption=false,font=normalsize,labelfont=sf,textfont=sf]{subfig}
\usepackage{textcomp}
\usepackage{stfloats}
\usepackage{url}
\usepackage{verbatim}
\usepackage{graphicx}
\usepackage{cite}
\usepackage{amssymb}
\usepackage{bm}
\usepackage{multirow}
\usepackage{booktabs}
\usepackage{bbding}
\usepackage{float}
\usepackage{placeins}
\usepackage{tabu}
\usepackage{makecell}
\usepackage{xcolor}
\definecolor{mycustompurple}{RGB}{154, 36, 79} % 定义自己的颜色
\usepackage[utf8]{inputenc}
\usepackage{hyperref} % 引入超链接宏包
\hypersetup{
    colorlinks=true,            % 激活链接颜色，去掉链接边框
    linkcolor=red,              % 文档内部链接颜色（如图表等引用）
    citecolor=green,            % 文献引用链接颜色
    filecolor=mycustompurple,   % 文件链接颜色
    urlcolor=magenta            % 外部URL链接颜色
}
\begin{document}

\title{MTSGL: Multi-Task Structure Guided Learning for Robust and Interpretable SAR Aircraft Recognition}

\author{Qishan He, Lingjun Zhao, Ru Luo, Siqian Zhang, Lin Lei, Kefeng Ji,~\IEEEmembership{Member,~IEEE,}, Gangyao Kuang,~\IEEEmembership{Senior Member,~IEEE,}
        % <-this % stops a space
\thanks{This paper was produced by the IEEE Publication Technology Group. They are in Piscataway, NJ.}% <-this % stops a space
\thanks{Manuscript received April 19, 2021; revised August 16, 2021.}}

% The paper headers
\markboth{Journal of \LaTeX\ Class Files,~Vol.~14, No.~8, August~2021}%
{Shell \MakeLowercase{\textit{et al.}}: A Sample Article Using IEEEtran.cls for IEEE Journals}

\maketitle

\begin{abstract}Aircraft recognition in synthetic aperture radar (SAR) imagery is a fundamental mission in both military and civilian applications. Recently deep learning (DL) has emerged a dominant paradigm for its explosive performance on extracting discriminative features. However, current classification algorithms focus primarily on learning decision hyperplane without enough comprehension on aircraft structural knowledge. Inspired by the fined aircraft annotation methods for optical remote sensing images (RSI), we first introduce a structure-based SAR aircraft annotations approach to provide structural and compositional supplement information. On this basis, we propose a multi-task structure guided learning (MTSGL) network for robust and interpretable SAR aircraft recognition. Besides the classification task, MTSGL includes a structural semantic awareness (SSA) module and a structural consistency regularization (SCR) module. The SSA is designed to capture structure semantic information, which is conducive to gain human-like comprehension of aircraft knowledge. The SCR helps maintain the geometric consistency between the aircraft structure in SAR imagery and the proposed annotation. In this process, the structural attribute can be disentangled in a geometrically meaningful manner. In conclusion, the MTSGL is presented with the expert-level aircraft prior knowledge and structure guided learning paradigm, aiming to comprehend the aircraft concept in a way analogous to the human cognitive process. Extensive experiments are conducted on a self-constructed multi-task SAR aircraft recognition dataset (MT-SARD) and the effective results illustrate the superiority of robustness and interpretation ability of the proposed MTSGL.
\end{abstract}

\begin{IEEEkeywords}Aircraft Recognition, Synthetic Aperture Radar (SAR), Multi-task Learning (MTL), Feature Disentangling, Stucture Consistency.
\end{IEEEkeywords}

\section{Introduction}
\IEEEPARstart{S}{ynthetic} aperture radar (SAR) automatic target recognition (ATR) is a fundamental component in the SAR image interpretation and plays an invaluable role in both military and civilian contexts \cite{RN1587}. Aircraft type classification is one of the meaningful tasks in SAR target recognition. In recent years, deep learning (DL) has witnessed an explosive performance in a variety of vision work \cite{RN1559,RN602,RN1588,RN1609,RN1610}. The idea of classification, a primitive task in computer vision (CV), is natural and applicable in SAR target recognition as well. Indeed, modern architectural neural networks supervised by cross-entropy loss have prevailed over many traditional handcrafted feature-based methods for SAR target recognition \cite{RN999,RN1044,RN1198,RN1289}. However, despite significant interest in this idea following the success of CV, an underlying key issue still remains whether neural networks comprehend the target to be identified in SAR images in a manner analogous to the objects in natural images. We attempt to answer this question from the following perspectives of data and annotation.

\begin{enumerate}
  \item{The data scale and diversity differ between CV datasets and SAR target datasets. The substantial quantity of training data is a principal factor contributing to the exceptional performance of neural networks in a range of vision-related tasks. In comparison, the SAR target dataset size is almost trivial due to the high acquisition cost and private ownership. The severe vulnerability to radar imaging conditions further raises huge cost of constructing the SAR target dataset covering enough diversity \cite{RN821}.}
  \item{Annotation messaging capacity is different between vision images and SAR images. Images in CV datasets are naturally occurring signals with heavy spatial redundancy \cite{RN1586}. A straightforward training strategy, e.g., supervision from categorical labels, appears to induce sophisticated image understanding. However, targets in SAR images often exhibit complex electromagnetic backscattering characteristics. Labelling in the simple form of categorical index hinders the experts' knowledge during the annotation operation, including prior awareness of specific targets' characteristics and conscious correlation between targets in SAR images and their inherent structures.}
\end{enumerate}

The above two limitations raise an urgent need to develop a learning-based algorithm with improved robustness and interpretation capabilities for practical SAR ATR. In this work, we take a different direction and develop a novel training paradigm based on a structure-based annotation for SAR aircraft recognition. The intuition is the following: the discriminative classification objective attends only to learning the hyperplane boundaries between the C classes on the D-dimensional feature spaces. The label, symbolized as a one-hot vector, reflects neither the individual characteristics nor the semantic concept of the targets \cite{RN1608}. \textcolor{black}{In this sense, it is imperative to explicitly exploit the expert knowledge about aircraft besides the one-hot label during the training stage.} The benefits from these unique tasks can then contribute to robust and interpretable recognition.

This powerful idea goes back to recent work on airplane recognition in optical remote sense images (RSI) \cite{RN1579,RN1580,RN1582,RN1581,RN1583}. Many researchers resorted to exploiting the structural and compositional annotation to facilitate the airplane recognition task. Chen et al. \cite{RN1579} proposed to infer the polygonal segmentation and used the overall shape to implement airplane recognition. Zuo et al. \cite{RN1580} introduced a coarse segmentation scheme that utilizes eight aircraft key-points, which provided a basic shape and scale information for airplane recognition. Qian et al. \cite{RN1582} and Jia et al. \cite{RN1581} proposed two component detection modules, where the aircraft wings and tail engines are extracted as dominant component features to enhance the classification performance. Zhao et al. \cite{RN1583} transformed aircraft recognition into the landmark regression task, which described the aircraft shape with several landmark points. It can be concluded that the annotation of global and local aircraft components has held the promise of enabling the neural network to understand the expert knowledge of aircraft, thereby yielding better recognition in optical RSIs. However, the explicit identification of the above elements, including landmarks, polygonal boundaries and components, is contingent upon the accurate contours of the aircraft and a pronounced distinction against the background. Aircraft in SAR images often exhibit complex electromagnetic backscattering characteristics \cite{RN1605}, including unclear target outlines and discrete appearance, impeding the availability of the aforementioned annotation methods.

Since aircraft have settled shapes in RSIs with appropriate geometric correction, we first introduce a structure-based annotation method, which is more suitable for aircraft in SAR imagery. Compared to the pixel-level label, the proposed annotation can obtain a structural segmentation mask based on pre-defined structure templates. The annotation is accomplished through the translation and rotation operation on the well-established image processing software, Photoshop, that ensures the operational convenience and speed. After that, a traditional geometric matching method is utilized to produce the affine deformation between the mask and the template. Overall, we obtain a three-fold annotation for each target chip, including category, mask and projection. The annotation incorporates the aircraft knowledge and conscious correlation between aircraft and the inherent structure in SAR images, thereby providing structural and compositional supplement information for deep learning.

Then, we propose a deep learning method for SAR aircraft recognition, named multi-task structure guided learning (MTSGL). Multi-task learning can handle different related tasks simultaneously in one shared system. With the relevance of different tasks, multi-dimensional information of target attributes is accessible to the neural network \cite{RN1589,RN1590}. The proposed method incorporates two structure guided tasks into the classification framework, i.e., structural semantic awareness (SSA) and structural consistency regularization (SCR). The core aim is to discover shared structure across tasks in a way that achieves greater efficiency and performance than solving the classification task individually. Our conjecture is that if the feature extracted from a classification network is exposed to predicting densely structural predictions, it should be able to comprehend the target knowledge. This can be succinctly articulated by Feynman’s mantra “What I cannot create, I do not understand”. Firstly, the SSA task provides a pixel-to-pixel objective, exposing the network to comprehensively understand aircraft concepts and capture the intrinsic class characteristics from discrete and sparse scattering points in SAR images. Secondly, the SCR task encourages learning geometrically consistent features corresponding to the affine-transformed templates. The structure consistency between features from the SAR image and the pre-defined template is maintained under the constraint of SCR. Thirdly, the classification task is employed to learn the decision boundary to complement the target recognition. At the testing stage, only the classification task is retained for keeping the end-to-end characteristic.

\begin{figure*}[!t]
  \centering
  \includegraphics[width=4.5in]{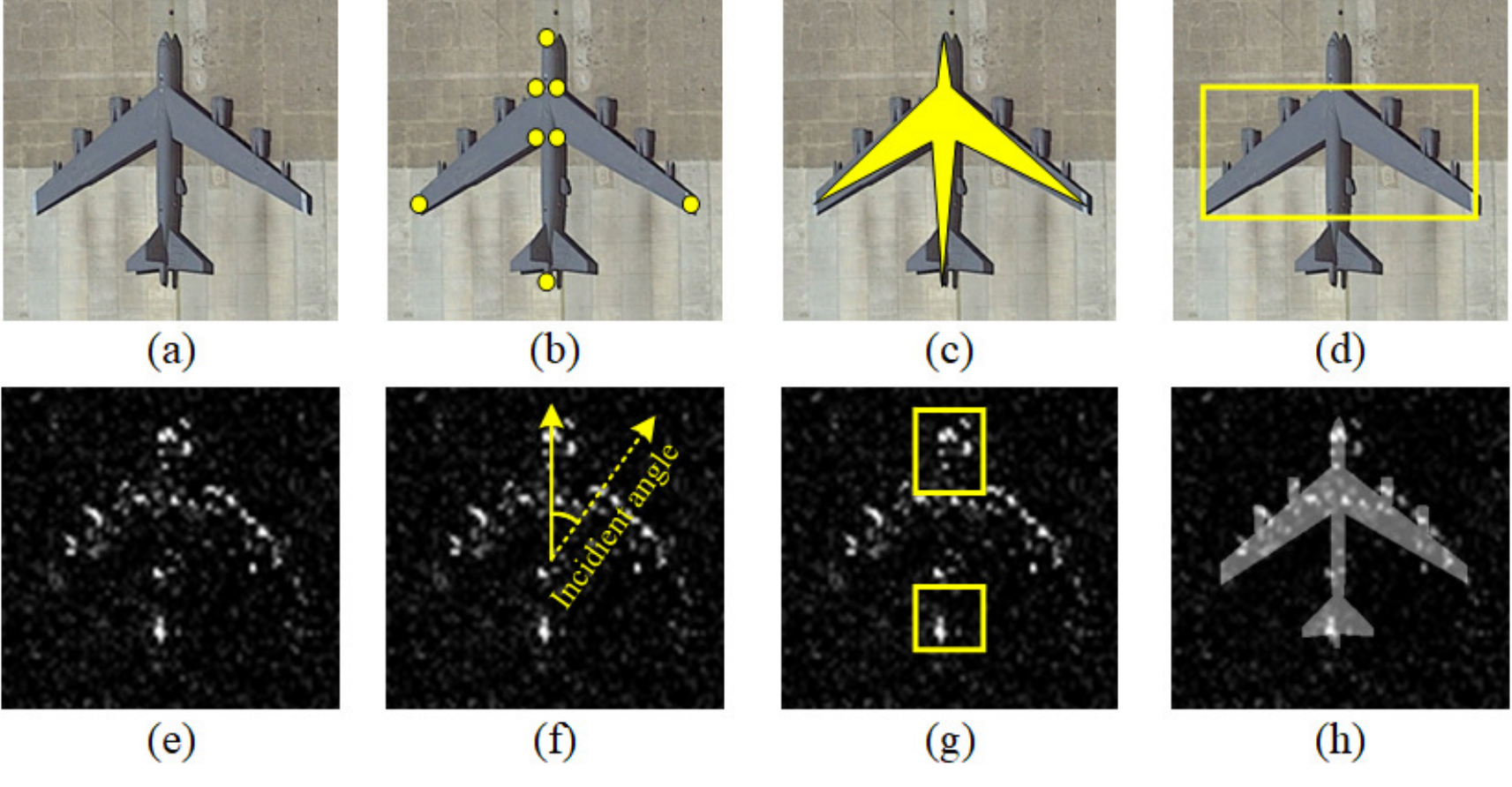}
  \caption{Illustration of existing aircraft annotation methods in the optical and SAR imagery. (a) Optical image. (b) Landmark-based \cite{RN1583}. (c) Polygon-based \cite{RN1580}. (d) Wing-based \cite{RN1582}. (e) SAR image. (f) Posture-based \cite{RN1585}. (g) Head and tail-based \cite{RN1591}. (h) Structure-based (Proposed).}
  \label{fig_1}
\end{figure*}

This article presents a multi-task learning framework for SAR aircraft recognition. The proposed framework aims to address the fundamental issue of improving the generalization and interpretation of induced deep features. The contributions of this article can be summarized as follows.
\begin{enumerate}
\item{We introduce a novel SAR aircraft annotation approach based on pre-defined structure templates. \textcolor{black}{Compared to the existing annotation approaches, first, it avoids refined annotation for pixel or boundary, hence guaranteeing operational convenience; second, it can induce equivalent information to that provided by other annotation methods, for instance, the attitude can be obtained from the rotation in the transformation matrix and the pixel-based interpretation can be derived from the geometrically transformed template.}}
\item{We propose a novel multi-task learning (MTL) framework MTSGL which incorporates the SSA and SCR tasks into classification training. The SSA disentangles the structure attributes by learning a pixel-to-pixel prediction objective. The SCR maintains the geometric consistency between targets in SAR image and templates. The induced feature will contain the basic aircraft structural knowledge, which essentially mimics the process of human comprehension of aircraft to increase the combinational generalization ability.}
\item{\textcolor{black}{The experiments on a self-constructed SAR aircraft recognition dataset demonstrate the superiority on robustness and interpretation of the proposed method. It gains a recognition accuracy increase of 6.84\%, 9.28\% and 10.49\% compared to the baseline at cases with 100\%, 70\% and 40\% training data respectively. The interpretability analysis demonstrates reliable decision evidence that offers users a more profound understanding of the recognized aircraft.}}
\end{enumerate}

The rest of this article is organized as follows. In Section II, we briefly introduce the related work. Section III explains the proposed structure-based annotation method and the details of MTSGL for aircraft recognition. In Section IV, the experiment results are presented and analyzed. Finally, Section V concludes this article.

\section{Related Work}
\label{section_II}
\subsection{Annotations of Aircraft Recognition}
Aircraft exhibits inerratic shapes and discriminative colors compared with backgrounds in optical RSIs, which provides crucial visually identification information. Many researchers focused their efforts on fined annotations to facilitate the airplane recognition task. Motivated by facial landmark detection, Zhao et al. \cite{RN1583} proposed to describe the aircraft with eight landmarks and trained a regression network to predict them. Consequently, the spatial geometric constraints among landmarks could be fully exploited to extract discriminative and reliable feature. In order to facilitate the distinction of ARJ21 from other civil airplanes, Jia et al. \cite{RN1581} marked the tail engine component and detected it using a faster R-CNN. The engine feature can reduce interference from airplanes of the other type with similar shape characteristics. Following the component discrimination feasibility, Qian et al. \cite{RN1582} used the oriented bounding box (OBB) to mark the wing component. The OBB detection result was utilized to fine-tune detection outcomes thus achieving enhanced performance. Chen et al. \cite{RN1579} introduced a semantic segmentation scheme where the airplane is outlined with polygonal boundaries. Zuo \cite{RN1580} et al. further employed octagonal segmentation that avoided the requirement of excessively refined annotation. By exploiting these annotations as the supervision, the deep neural network can extract more reliable features which contain rich structural and componential information.

For aircraft in SAR images, Kang et al. \cite{RN1585} supplemented the target attitude angles to the annotations in addition to the categories. He et al. \cite{RN1591} labeled the aircraft's head and tail as component information in SAR imagery. Although the above labelling methods are able to characterize the attitude and components of the aircraft in SAR imagery as shown in Fig. \ref{fig_1}, there is still a lack of labelling techniques to characterize the structure of the aircraft in the same way as for aircraft in optical imagery. It is challenging to delineate boundaries and landmark points at pixel-level due to the discrete and fuzzy aircraft appearance in SAR images.

\begin{figure*}
\centering
\includegraphics[width=7in]{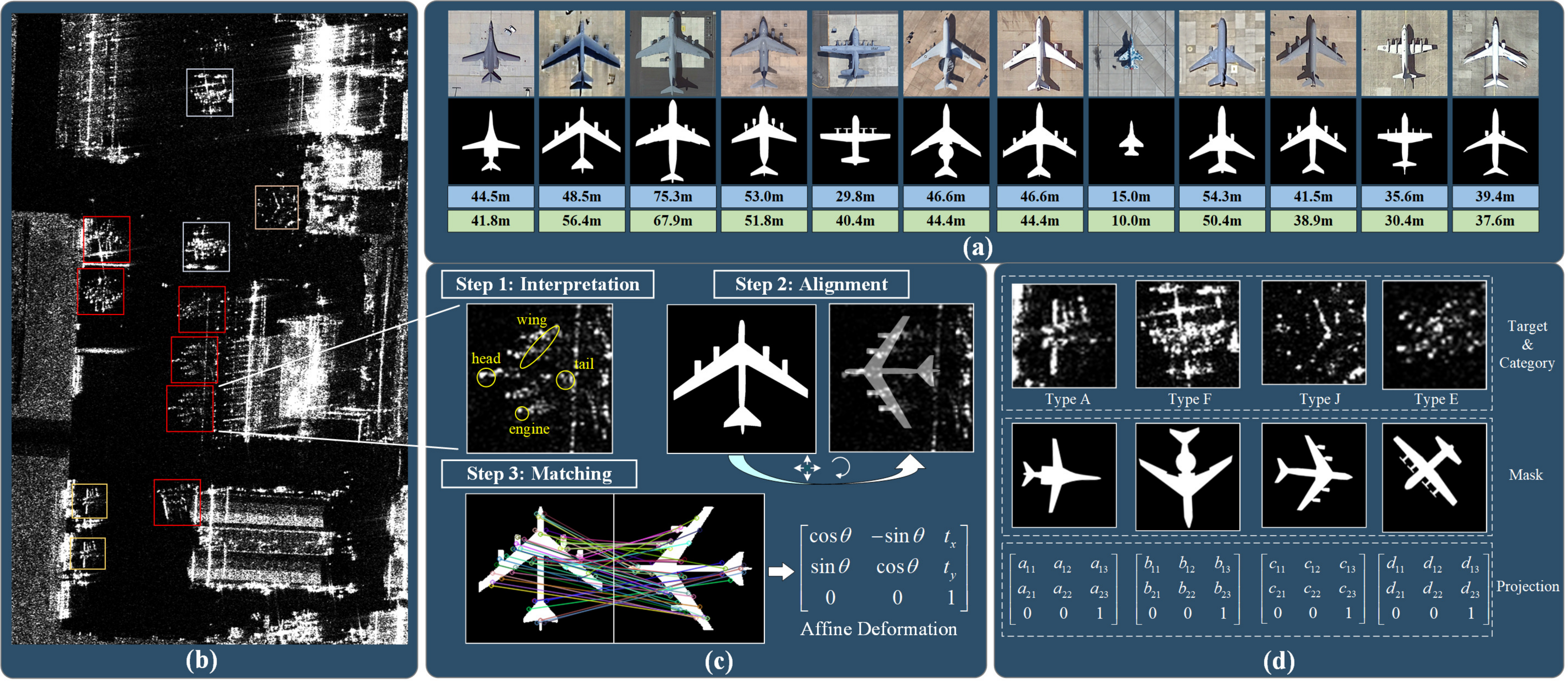}
\caption{Overall description of the proposed template-based annotation for SAR aircraft target. (a) Predefined binary templates and scale information of aircraft with lengths in blue blocks and wingspans in green blocks. (b) Initial annotation based on horizontal bounding boxes in panoramic images. (c) Expert-level annotation on target area. This process consists of two manual steps and one automated step, including interpretation, alignment and matching. (d) The formed SAR aircraft recognition dataset, involving origin images, categories, masks and projections.}
\label{fig_2}
\end{figure*}

\subsection{DL based SAR Airplane Recognition}
In recent years, SAR ATR’s researchers have conducted a variety of deep learning algorithms, that promoted substantial improvements in recognition performance on the MSTAR vehicle dataset \cite{RN1195,RN1177,RN1054,RN1433}. Compared with the traditional methods of manually designing features, the CNN-based methods reduce the parameters and provide a strong feature representation ability. However, they lack the consideration with SAR target characteristics and have poor robustness of feature representation. For fined SAR airplane classification, Kang et al. \cite{RN1585} utilized the scattering characteristics of aircraft and integrates the spatial topology feature into classification branch to predict the class scores. Sun et al. \cite{RN1518} designed a few-shot learning framework where the strong scattering points are predicted as an optimization assistance for the main classification. Lv et al. \cite{RN1592} designed a scattering correlation classifier incorporated with a quantitative model characterizing the discretization degree of aircraft targets. Zhao et al. \cite{RN1593} employed a Gaussian mixture distribution model (GMM) to ascertain the spatial distribution of strong scattering points. Then a graph neural network \cite{RN1604} is designed to extract point-level features. 

The current dedicated approaches primarily focused on analyzing the scattering characteristic of SAR aircraft to get better perception of the aircraft topology. \textcolor{black}{Most research involved another hand-crafted feature branch, which complicated the algorithmic process and worked against the end-to-end characteristic \cite{RN1585, RN1518, RN1592, RN1593}}. Moreover, the SAR aircraft recognition is typically treated as object classification task, whereby only the most distinctive features are retained with less emphasis placed on the aircraft structural knowledge. What we need is a principled approach that allows the network to perceive the aircraft structural knowledge autonomously.

\section{Proposed Method}
\label{section_III}
In this section, we first introduce the proposed structure-based annotation method for SAR aircraft. On this basis, a novel multi-task learning framework named MTSGL is then proposed. The framework integrates two structure guided tasks, structural semantic awareness and structural consistency regularization, into the transformed-based classification network. An adaptive multi-gradient descents algorithm based on Pareto optimization is introduced to produce the globally optimal solution.

\subsection{Structure-based SAR Aircraft Annotation}
We firstly proposed a SAR aircraft annotation method based on pre-defined structure templates, as illustrated in Fig. \ref{fig_2}. The shape and color context of aircraft in optical RSIs are sufficiently distinctive to enable the generation of a binary segmentation ground truth template for each aircraft class, as shown in Fig. \ref{fig_2}(a). \textcolor{black}{The annotation for all 12 aircraft templates is implemented using the LabelMe software. This process can be regarded as the extraction of prior knowledge from optical images.} Then, the aircraft in SAR images are labeled according to the corresponding optical images and expert knowledge. As show in Fig. \ref{fig_2}(b), different colors represent different categories.

While aircraft in SAR images typically manifest a discontinuous and incomplete appearance, there are often discernible scattering points specific to individual components, which can facilitate the manual identification of specific aircraft categories. For example, an expert annotation operator could infer the structural shape through analyzing the edge diffraction of the wing and reflection points of the fuselage. The length of the aircraft can be determined by measuring the distance between the scattering points at the front and back. The presence of a cavity structure within the engine component results in the generation of strong backscattering, which serves to confirm the locations and counts of the aforementioned components. In light of the aforementioned analysis, our objective is to facilitate the neural network's capacity to discern the intrinsic aircraft structure from the SAR image in a manner analogous to that of an expert interpreter.

\textcolor{black}{We resort to the predefined template to interpret the SAR aircraft image, which induces a segmentation ground truth and a geometry affine deformation. The step 2 is to manually align the pre-defined templates with the SAR aircraft image slices.} This is accomplished in the well-established image processing software, Photoshop, which provides translation and rotation operation across different layers, as shown in Fig. \ref{fig_2}(c). It should be emphasized that the pixel spacing of the template image is adjusted to match the actual spacing in the SAR images, so no scaling operation is required. Consequently, the aligned templates are exported as segmentation solutions.

\begin{figure*}
\centering
\includegraphics[width=7in]{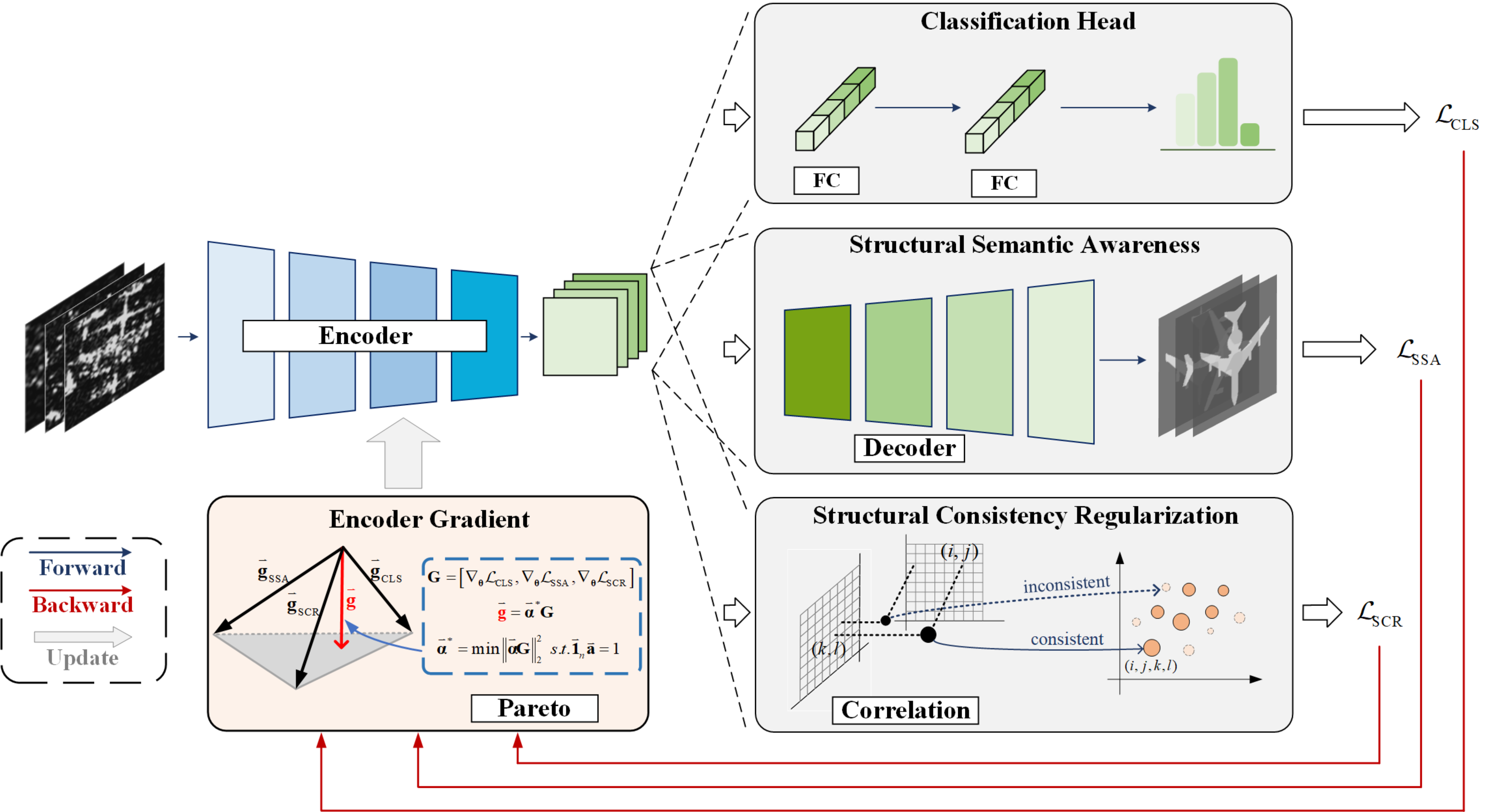}
\caption{Overall illustration of the proposed MTSGL. The multi-objectives are obtained by passing the input images through one shared feature encoder and three task-specific modules. Consequently, the encoder's gradient is obtained by merging three task-specific gradients backpropagated from their respective losses. At testing phase, only the classification branch is preserved to predict the class scores.}
\label{fig_3}
\end{figure*}

\textcolor{black}{We further conduct geometric matching between the aligned mask and template to record the manual translation and rotation magnitudes, which represented by the step 3 in Fig. \ref{fig_2}(c). This step obtains a geometric deformation matrix.} \textcolor{black}{Oriented fast and rotated brief (ORB) features \cite{RN1594} are extracted from both images and then matched based on the Brute Force algorithm with k-nearest neighbor searching (KNN). k is set to 2. When comparing the distance ratio between the two best matches found by KNN, and if the ratio between the two distance is smaller the threshold, the first is selected. The distance ratio threshold is set to 0.75.} The selected feature pairs are utilized to estimate the affine transformation, which is a 6 degree-of-freedom linear transformation capable of modeling manual alignment operation. It can be parameterized by a 6-dimensional vector
\begin{equation}
\label{eq1}
M=\begin{bmatrix}\cos\theta&-\sin\theta&t_x\\\sin\theta&\cos\theta&t_y\end{bmatrix}\overset{\Delta}{=}\begin{bmatrix}a_{11}&a_{12}&a_{13}\\a_{21}&a_{22}&a_{23}\end{bmatrix}
\end{equation}
such that points $P_B=\begin{bmatrix}x_B,y_B\end{bmatrix}^\text{T}$ are transformed to points $P_A=\begin{bmatrix}x_A,y_A\end{bmatrix}^\text{T}$ according to the mapping function
\begin{equation}
\label{eq2}
\mathcal{T}_M\left(P_B\right)=\begin{bmatrix}a_{11}&a_{12}\\a_{21}&a_{22}\end{bmatrix}P_B+\begin{bmatrix}a_{13}\\a_{23}\end{bmatrix}.
\end{equation}

Obviously, there are several fundamental differences between the proposed method and existing annotation methods. First, the pixel-level marking for boundaries and points is substituted by manipulating the layer-layer relationship in Photoshop. This guarantees the operational convenience and speed. Second, our method offers a global aircraft interpretation in pixel level and a geometry correspondence with respect to the structure template, which we believe are the key factors for the neural network understanding of aircraft targets in SAR images.

\textcolor{black}{It is indisputable that this concept does exacerbate the requirement for human expertise. In the context of conventional object classification in computer vision, such annotation is not required except for category information. Hence, the research will continue in the future, with the aim of estimating the geometry transformation $M$ between the SAR aircraft and the template. The automated acquisition of such high-level information is considered the ultimate objective, given its practical utility in aircraft recognition.}

\subsection{Multi-Task Structure Guided Learning (MTSGL)}
Multi-task learning is a powerful tool for learning multiple correlated tasks at the same time. With the relevance of the different tasks, neural networks can achieve greater performance of each task than training individually \cite{RN1596}. This article generalizes this idea on SAR aircraft recognition for developing an expert-like comprehension of neural networks. We firstly present the concept of conducting the SSA task and the SCR task in addition to the classification task.

It should be borne in mind that, during the proposed annotation process, a human specialist must analyze the discrete appearance of SAR aircraft and manually align the targets with the predefined templates, as shown in the interpretation and alignment step of Fig. \ref{fig_2}(c). These steps provide a firmer judgement associated with component and structure for identifying aircraft's types. Our objective is to mimic this process using various underlying vision tasks. First, pixel-to-pixel prediction, trained with pixel-wise loss, is able to predict a high-dimensional structured outputs. A thorough comprehension of the detailed semantics of the constituent components is a necessary prerequisite for succeeding at this undertaking. \textcolor{black}{Second, structural consistency between the SAR image and template image should be maintained to obtain consistent and stable features for aircrafts with the same category but different orientations. The SCR task imposes spatial constraints among aircraft parts for model optimization which is particularly beneficial for robust recognition, as aircraft are typically highly sensitive to azimuthal variance in SAR images. By iteratively restricting the SCR loss driven the deformation matrix, a structure-consistent model against geometry variance can be guaranteed.}

Accordingly, the proposed multi-task framework consists of one shared encoder and three task-specific heads, as shown in Fig. \ref{fig_3}. The encoder is a classical hierarchical transformer \cite{RN1601}, which has yielded encouraging outcomes in vision tasks, specifically image classification. It inputs images with size of $H\times W$ and produces hierarchical representations. The final feature is usually with the size of $h\times w\times d$, where $h=\frac{H}{32}$, $w=\frac{W}{32}$. The global average pool (GAP), fully connection (FC) layer and Softmax function are applied to get classification probabilities. In terms of the classification loss, the cross entropy is used for evaluating the distance between predicted logits and the true labels, as following
\begin{equation}
\label{eq3}
\mathcal{L}_{\mathrm{CLS}}=-\sum_{c=1}^{C}y_{c}\log(\hat{p}_{c}).
\end{equation}
where $C$ is the number of categories. 

\subsubsection{Structural Semantic Awareness (SSA)}

\begin{figure*}[!t]
\centering
\includegraphics[width=6in]{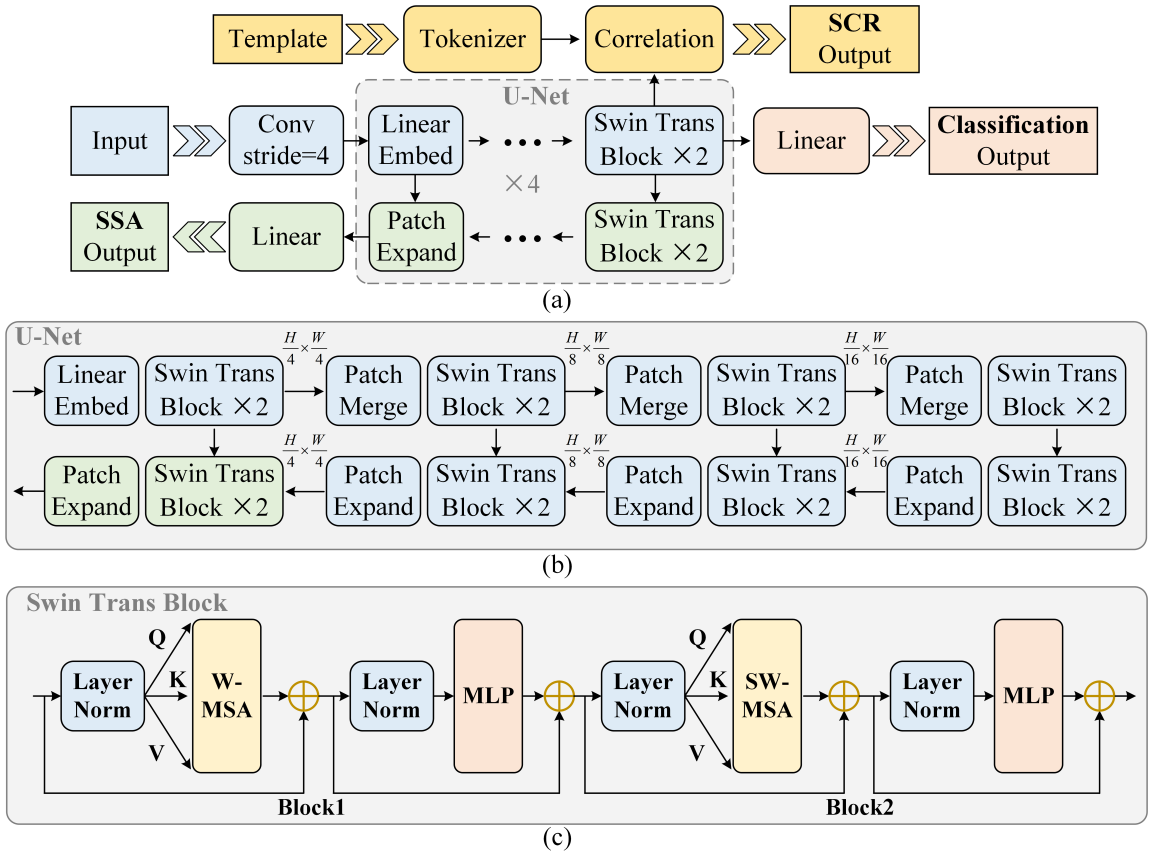}
\caption{\textcolor{black}{The transformer-based structure of the MTSGL. (a). The forward process given the input image and the corresponding template during the training stage. Note that the SCR branch (yellow) are discarded at inference stage. The feature encoder (blue) and the SSA branch (green) adopt the U-Net structure. (b). The U-Net detailed structure. (c). The Swin Transformer block.}}
\label{fig_4}
\end{figure*}

\begin{figure}[!t]
  \centering
  \includegraphics[width=3.5in]{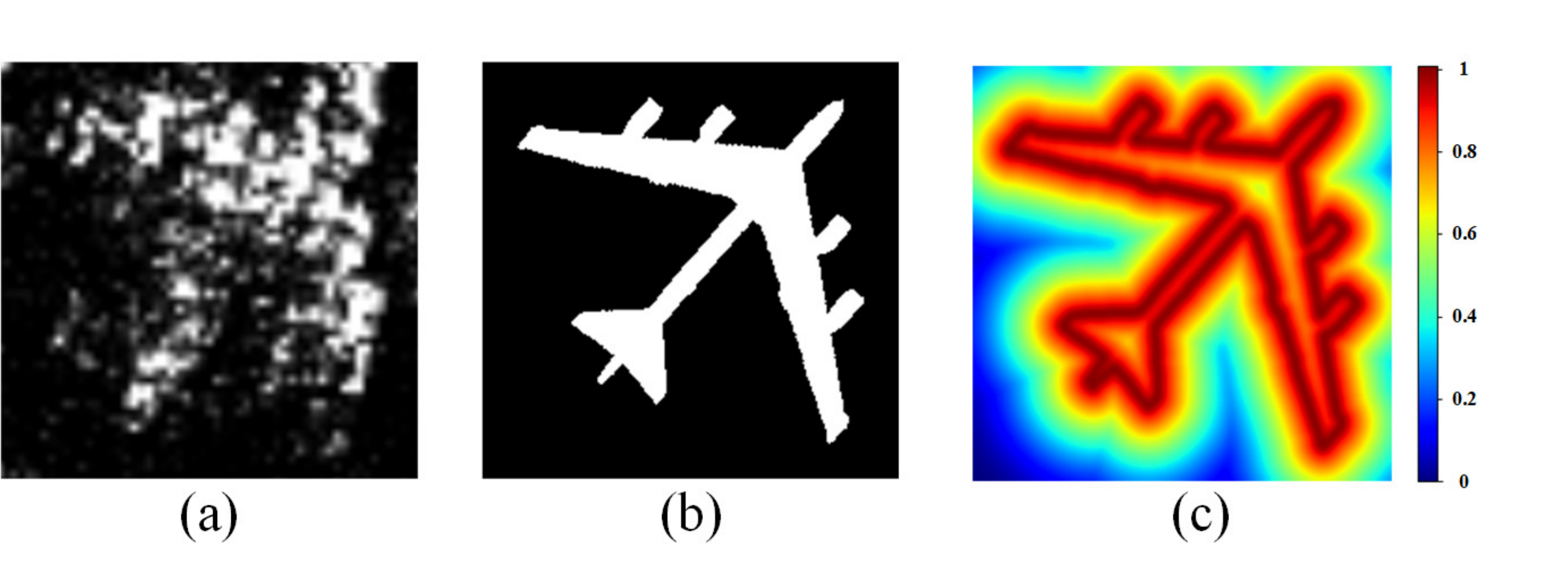}
  \caption{Illustration of the distance map from the segmentation ground truth. (a) SAR image. (b) Ground Truth. (c) Distance Map.}
  \label{fig_16}
\end{figure}

Structural semantic awareness is achieved in a pixel-to-pixel prediction task, which enables the encoder to extract features with comprehensive structural information. \textcolor{black}{To this end, a symmetric transformer-based encoder-decoder structure following U-Net is built as shown in Fig. \ref{fig_4}(b). The basic unit is Swin Transformer block. For the encoder, it transforms the input into a feature map with spatial size of $\frac{H}{4}\times\frac{W}{4}$ via a convolutional layer with the stride set to 4. Then it is transformed into tokens by linear embedding. In the encoder of U-Net, multi-scale hierarchical features are obtained through several patch merging layers that perform 2x down-sampling of resolution. The decoder is composed of Swin Transformer block and patch expanding layer.} 

Specially, the skip-connection is used to fuse the multi-scale features from the decoder with the up-sampled features. The up-sampling is achieved by patch expanding, which is an inverse operation of the patch merging layer. Patch expanding rearranges the input features to twice their original resolution while simultaneously reducing the channel number to half. The feature tokens from the last block is reshaped and linearly projected to the original resolution with $C+1$ channels. Consequently, the output is then subjected to the Softmax function, which yields probabilities across the C aircraft categories and one background category. 

Typical pixel level cross entropy works well as a distribution difference measure for segmentation objective. However, it equally treats each pixel sample during training, which is not sufficient to perceive structure information. The ground truth is an aligned aircraft template binary image, where contours and small-scale components are important for tracking the aircraft's shape and structure. Following \cite{RN1612}, we introduce distance maps to guide the network to focus towards boundary regions, thereby obtaining shape-enhanced structure semantic awareness. On this basis, the SSA loss is calculated by

\begin{equation}
    \label{eq3_}
    \mathcal{L}_{\mathrm{SSA}}=-\sum_{i=1}^N(1+\Phi)\odot\sum_{c=1}^Cy_{ic}\log(\hat{y}_{ic})
\end{equation}
where the two sums run over the $i$-th pixel sample and the $c$-th class, and $\odot$ is the Hadamard product. $\Phi$ is the distance weight term generated by taking the inverse of the distance transform of ground truth \cite{RN1612}, as shown in Fig. \ref{fig_16}. In this way, the pixels in proximity of the boundary are weighted more, compared to those located far away.

The U-Net structure is important for our multi-task learning framework. This can be explained by the gap between the structure prediction task and the classification task. The last several layers in the encoder are more specialized to compress the feature dimension for classification, but less pertinent to generating the high-dimension structural output. The shortcuts since the encoder and decoder can alleviate the pressure to learn structure prediction representations, because low-level detailed features are directly fed to the decoder. By this mean, the shared encoder can focus on learning discriminative features while maintaining high-level structural semantic awareness capability.

\subsubsection{Structural Consistency Regularization (SCR)}
The structural consistency between the SAR image chip and the pre-defined template is the key factor to implement the proposed aircraft annotation method. It should be maintained throughout the learning process during model training. Our hypothesis is that the structural consistency regularization enables the encoder to better comprehend the fundamental structural characteristics of aircraft during SSA.

\begin{figure}[!t]
\centering
\includegraphics[width=3.3in]{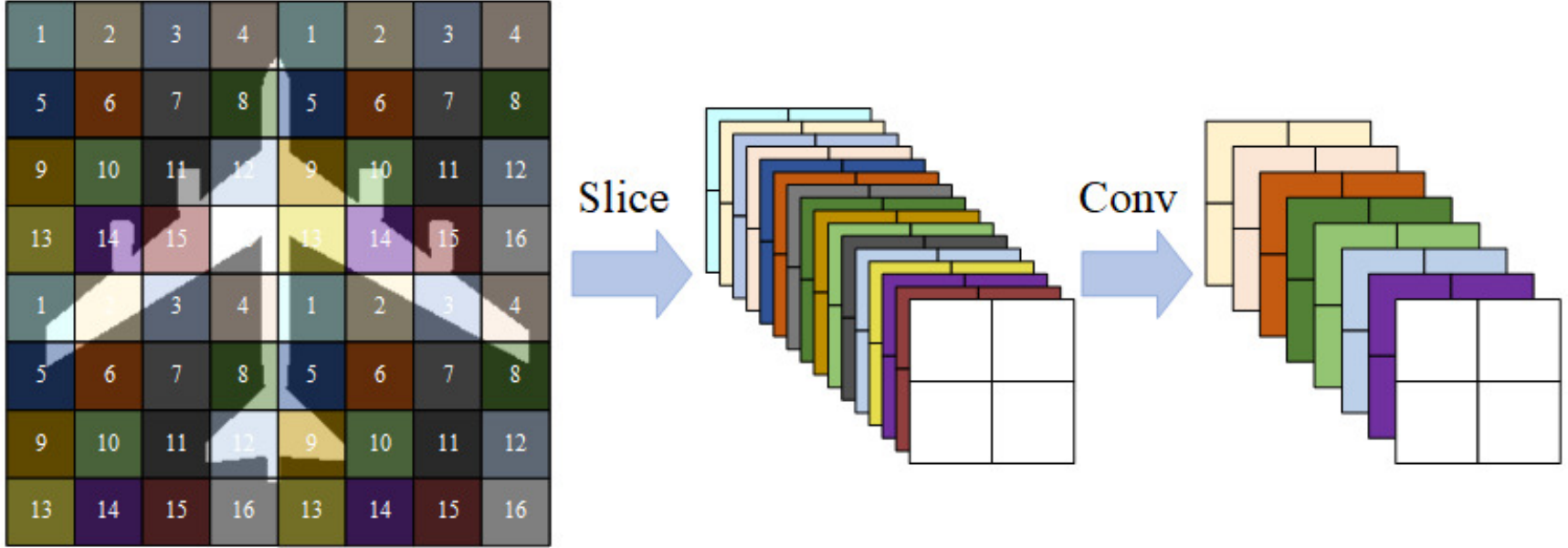}
\caption{Structure template tokenizer module.}
\label{fig_5}
\end{figure}

Besides the transformer encoder, an independent feature tokenizer branch is additionally designed to encode the pre-defined structure template into tokens with equal size of the encoder's output. As shown in Fig. \ref{fig_5}, this branch firstly slices the template image into patches with size of $N\times N$ and flattens the $N^{2}$ pixels in each patch into channel dimension. A $3\times 3$ convolution padded with 1 is then performed to align the channel number between the outputs from the encoder and the tokenizer module. It can be observed that the designed tokenizer branch does not loss any spatial structure information, that is, the relative positional relationship of pixels on the initial image and the feature token remains unaltered.

The input SAR image and the pre-defined structure template, $\begin{pmatrix}I_s,I_t\end{pmatrix}$, are passed through the encoder and the designed tokenizer module respectively. The resulting feature maps $\begin{pmatrix}f_{s},f_{t}\end{pmatrix}$ are two $h\times w\times d$ tensors which can be interpreted as two $h\times w$ grids of $d$-dimensional tokens, i.e., $f_{s}^{i,j},f_{t}^{k,l}\in\mathbb{R}^{d}$. We apply a normalized correlation layer to $\begin{pmatrix}f_{s},f_{t}\end{pmatrix}$ and obtain the correlation map $C_{s,t}\in\mathbb{R}^{h\times w\times h\times w}$, calculated as
\begin{equation}
    \label{eq6}
    C_{s,t}^{ijkl}=\frac{\left\langle f_{s}^{i,j},f_{t}^{k,l}\right\rangle}{\sqrt{\sum_{a,b}\left\langle f_{s}^{a,b},f_{t}^{k,l}\right\rangle}}.
\end{equation}
where the 4-D element $C_{s,t}^{ijkl}$ records the normalized inner product at two locations $\begin{bmatrix}i,j\end{bmatrix}$ in $f_{s}$ and $\begin{bmatrix}k,l\end{bmatrix}$ in $f_{t}$. The resulting $C$ contains the match scores of all token pairs between the SAR images and the structure template. Similarly, the correlation map in the opposite direction is also considered as
\begin{equation}
    \label{eq6_}
    C_{t,s}^{klij}=\frac{\left\langle f_{t}^{k,l},f_{s}^{i,j}\right\rangle}{\sqrt{\sum_{a,b}\left\langle f_{t}^{a,b},f_{s}^{i,j}\right\rangle}}.
\end{equation}

\begin{figure}[!t]
\centering
\includegraphics[width=3.4in]{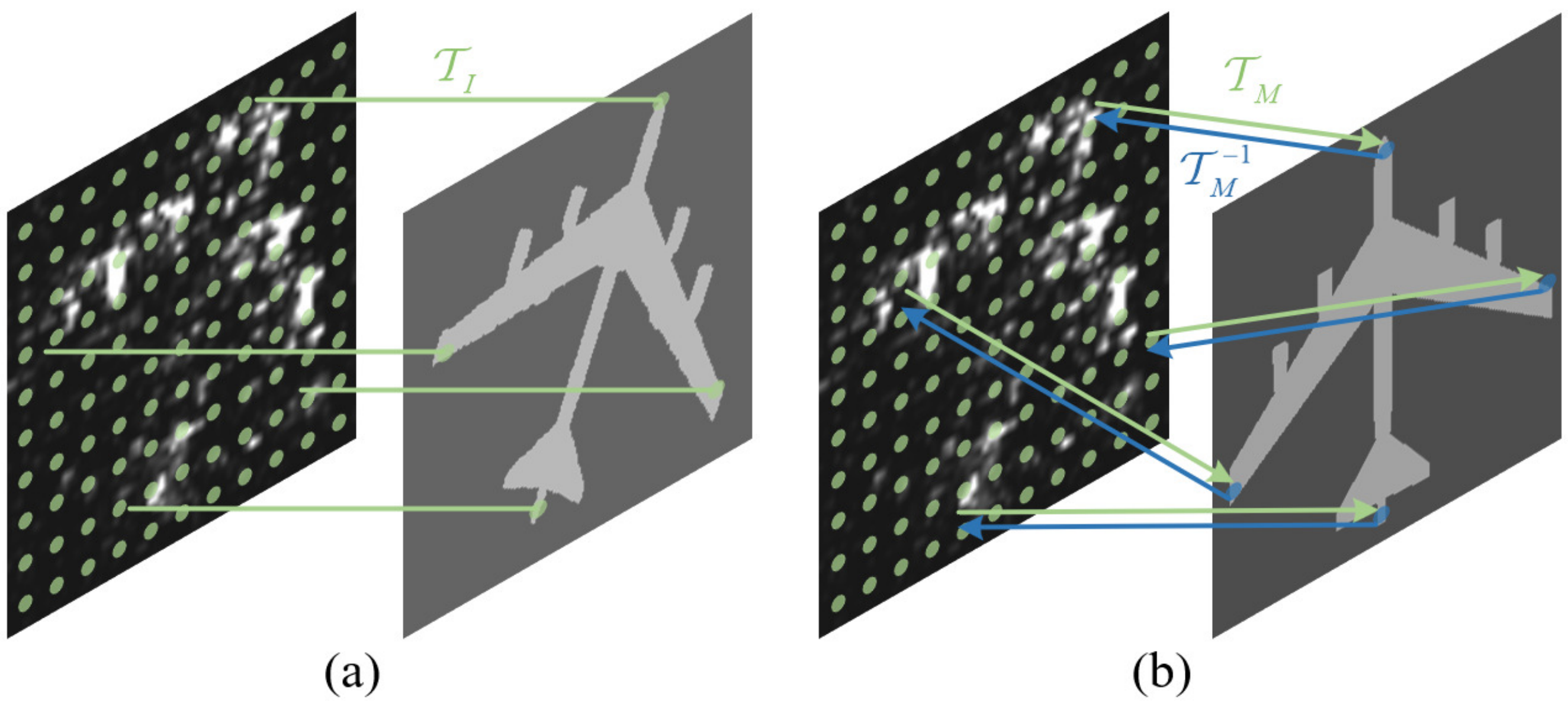}
\caption{Two examples of geometrically consistent correspondence. (a) $\mathcal{T}_I$ is an identity transformation. (b) $\mathcal{T}_M$ is an affine transformation.}
\label{fig_6}
\end{figure}

Given the affine transformation ground truth $M$, we can identify the token pairs with geometric consistency. It is obtained by two corresponding masks 
\begin{equation}
    \label{eq7}
    m_{s,t}^{ijkl} = \begin{cases}
    1,&{\text{if}}\ \mathrm{d}\left(p,\mathcal{T}_M\left(q\right)\right)\leq\varphi\\ 
    {0,}&{\text{otherwise.}} 
    \end{cases}
\end{equation}

\begin{equation}
    \label{eq7_}
    m_{t,s}^{klij} = \begin{cases}
    1,&{\text{if}}\ \mathrm{d}\left(q,\mathcal{T}_M^{-1}\left(p\right)\right)\leq\varphi\\ 
    {0,}&{\text{otherwise.}} 
    \end{cases}
\end{equation}
where $p=\begin{bmatrix}i,j\end{bmatrix}^T$, $q=\begin{bmatrix}k,l\end{bmatrix}^T$, $d(\cdot,\cdot)$ is distance function and $\varphi $ is a pre-defined threshold set to 1. The corresponding mask can be represented by a wrapped 4-D identity matrix using a spatial grid transformer layer $\mathcal{T}_M$ \cite{RN1598}. The effect of this transformation on the grid compared to the identity transform is shown in Fig. \ref{fig_6}.

The training objective is to maximize the correlation of the pairs with geometric consensus, thereby entangling the structural attribute in a geometrically meaningful manner. This objective would also penalize ambiguous correlations with inconsistent correspondence due to the normalization operation when calculating the correlation map. Lastly, the structural consistency regularization loss is calculated by summing the masked correlation scores in both directions, as follows
\begin{equation}
\label{eq8}
\mathcal{L}_{\mathrm{SCR}}=-\sum_{i,j,k,l}m_{s,t}^{ijkl}C_{s,t}^{ijkl}-\sum_{k,l,i,j}m_{t,s}^{klij}C_{t,s}^{klij}.
\end{equation}

\subsubsection{Pareto-Oriented Multi-Gradient Descent}
For the backward process, an intuitively appealing method is combining three tasks into one proxy objective via a linear weighted summation. Then the encoder's weights are updated by
\begin{equation}
\label{eq9}
\begin{split}
\bm{\theta}^{t+1}&=\bm{\theta}^{t}-\nabla_{\bm{\theta}}\left(w_{1}\mathcal{L}_\mathrm{CLS}+w_{2}\mathcal{L}_\mathrm{SSA}+w_{3}\mathcal{L}_\mathrm{SCR}\right)\\
&\triangleq\bm{\theta}^{t}-\nabla_{\bm{\theta}}\left(w_{1}\mathcal{L}_{1}+w_{2}\mathcal{L}_{2}+w_{3}\mathcal{L}_{3}\right).
\end{split}
\end{equation}

The drawback is obvious because determining appropriate weights requires an expensive grid search. More importantly, no single solution of the encoder can achieve the best performance on all tasks due to the gradient conflict caused by the inherent task differences. To avoid the biased update direction caused by \eqref{eq9}, we resort to achieve Pareto optimality which yields a global multi-objective solution \cite{RN1599,RN1600}. A solution $\bm{\theta}^{*}$ is called Pareto optimality if there exists no solution $\bm{\theta}$ at more dominant case where $\mathcal{L}_{i}(\bm{\theta})\leq\mathcal{L}_{i}(\bm{\theta}^{*})$, $\forall i\in\begin{Bmatrix}1,2,3\end{Bmatrix}$ and $\mathcal{L}_{j}(\bm{\theta})\leq\mathcal{L}_{j}(\bm{\theta}^{*})$, $\exists j\in\begin{Bmatrix}1,2,3\end{Bmatrix}$.

To derive the solution, we construct the constrained minimization problem
\begin{equation}
\label{eq10}
\min_{a_{1},a_{2},a_{3}}\left\|a_{1}\nabla_{\bm{\theta}}\mathcal{L}_{1}+a_{2}\nabla_{\bm{\theta}}\mathcal{L}_{2}+a_{3}\nabla_{\bm{\theta}}\mathcal{L}_{3}\right\|_{2}^{2}
\end{equation}
\begin{equation}
\label{eq10_}
s.t.\sum_{i=1}^{3}a_{i}=1,a_{i}\geq0,\forall i.
\end{equation}

The solution of this optimization problem, called Pareto stationary point, gives a descent direction that improves all tasks. In our setting, the number of tasks is $n=3$ and the dimension is the number of encoder’s parameters that can be millions. Therefore, we attempt to solve the equation \eqref{eq10} as a convex quadratic problem with linear constraints.

We use Lagrange multiplier method to convert the convex optimization into
\begin{equation}
\label{eq11}
\begin{aligned}
\mathcal{J}=\left\|\boldsymbol{a}\left[\nabla_{\bm{\theta}}\mathcal{L}_{1},\nabla_{\bm{\theta}}\mathcal{L}_{2},\nabla_{\bm{\theta}}\mathcal{L}_{3}\right]^{T}\right\|_{2}^{2}-\lambda\left(\mathbf{1}_{n}\boldsymbol{a}^{T}-1\right)
\end{aligned}
\end{equation}
where $\boldsymbol{a}=\begin{bmatrix}a_1,a_2,a_3\end{bmatrix}$ and $\mathbf{1}_n$ is a row vector with all elements equal to 1. By solving the partial differential equation of the Lagrangian, the resulting analytical solution is obtained as following
\begin{equation}
\label{eq12}
\boldsymbol{a}^{*}=\left[\begin{bmatrix}\mathbf{0}_{n}&1\end{bmatrix}\times\begin{bmatrix}\mathbf{G}&\mathbf{1}_{n}^{T}\\\mathbf{1}_{n}&0\end{bmatrix}^{-1}\right]_{1:n}
\end{equation}
where $\mathbf{0}_n$ is a row vector with all elements equal to 0, $\left[\cdot\right]_{1:n}$ represents taking the first $n$ elements and
\begin{equation}
\label{eq13}
\mathbf{G}=\begin{bmatrix}\left\|\nabla_{\bm{\theta}}\mathcal{L}_1\right\|_2^2&\left\langle\nabla_{\bm{\theta}}\mathcal{L}_1,\nabla_{\bm{\theta}}\mathcal{L}_2\right\rangle&\left\langle\nabla_{\bm{\theta}}\mathcal{L}_1,\nabla_{\bm{\theta}}\mathcal{L}_3\right\rangle\\\left\langle\nabla_{\bm{\theta}}\mathcal{L}_2,\nabla_{\bm{\theta}}\mathcal{L}_1\right\rangle&\left\|\nabla_{\bm{\theta}}\mathcal{L}_2\right\|_2^2&\left\langle\nabla_{\bm{\theta}}\mathcal{L}_2,\nabla_{\bm{\theta}}\mathcal{L}_3\right\rangle\\\left\langle\nabla_{\bm{\theta}}\mathcal{L}_3,\nabla_{\bm{\theta}}\mathcal{L}_1\right\rangle&\left\langle\nabla_{\bm{\theta}}\mathcal{L}_3,\nabla_{\bm{\theta}}\mathcal{L}_2\right\rangle&\left\|\nabla_{\bm{\theta}}\mathcal{L}_3\right\|_2^2\end{bmatrix}.
\end{equation}
The encoder is updated based on a reweighed summation by
\begin{equation}
\label{eq14}
\bm{\theta}^{t+1}=\bm{\theta}^{t}-\nabla_{\bm{\theta}}\left(\boldsymbol{a}^{*}\left[\nabla_{\bm{\theta}}\mathcal{L}_\mathrm{CLS},\nabla_{\bm{\theta}}\mathcal{L}_\mathrm{SSA},\nabla_{\bm{\theta}}\mathcal{L}_\mathrm{SCR}\right]^{T}\right).
\end{equation}

The geometric meaning of the updated gradient is equivalent to the perpendicular vector from the origin to the plane formed by the convex hull of the three task-specific gradient vectors \cite{RN1606}, as shown in Fig. \ref{fig_3}. The Pareto-oriented multi-objective optimization avoids the gradient conflict and adaptively ensures the global optimality.

{\textcolor{black}The training and inference procedures are presented in Algorithm \ref{algorithm1}, \ref{algorithm2}. The inference process remains the end-to-end characteristic that provides a classification and a segmentation prediction. The latter output supplies an intuitive classification decision evidence, which will be discussed in Section IV \ref{Section_IV_D}.
\begin{algorithm}
    \caption{MTSGL Training}\label{algorithm1}
    \KwData{Training Data $\mathcal{D}$}
    % \KwResult{Encoder $\bm{\theta}$, Three Task-specific Modules $\bm{\theta}_\mathrm{CLS},\bm{\theta}_\mathrm{SSA},\bm{\theta}_\mathrm{SCR}$}
    Initialization Encoder $\bm{\theta}$, Three Task-specific Modules $\bm{\theta}_\mathrm{CLS}$, $\bm{\theta}_\mathrm{SSA}$, $\bm{\theta}_\mathrm{SCR}$\;
    \While{converaged}{
    ${I^s}$, ${I^t}$, $M$ $\leftarrow$ Sample from $\mathcal{D}$ (or mini-batch)\;
    $f^s$, $f^t$ $\leftarrow$ Extract tokens from ${I^s}$, ${I^t}$\;
    $\mathcal{L}_{\mathrm{CLS}}$ $\leftarrow$ Calculate classification loss according to Eq. \ref{eq3}\;
    $\mathcal{L}_{\mathrm{SSA}}$ $\leftarrow$ Calculate SSA loss according to Eq. \ref{eq3_}\;
    $\mathcal{L}_{\mathrm{SCR}}$ $\leftarrow$ Calculate SCR loss according to Eq. \ref{eq8}\;
    $\boldsymbol{a}^{*}$ $\leftarrow$ Calculate the Pareto solution through Eq. \ref{eq12}\;
    Update the encoder's parameters $\bm{\theta}$ by $\boldsymbol{a}^{*}$-weighted summation of $[\nabla_{\bm{\theta}}\mathcal{L}_{\mathrm{CLS}}, \nabla_{\bm{\theta}}\mathcal{L}_{\mathrm{SSA}}, \nabla_{\bm{\theta}}\mathcal{L}_{\mathrm{SCR}}]$\;
    Update three task-specific modules' parameters $\bm{\theta}_\mathrm{CLS},\bm{\theta}_\mathrm{SSA},\bm{\theta}_\mathrm{SCR}$ by $\nabla_{\bm{\theta}}\mathcal{L}_{\mathrm{CLS}}$, $\nabla_{\bm{\theta}}\mathcal{L}_{\mathrm{SSA}}$ and $\nabla_{\bm{\theta}}\mathcal{L}_{\mathrm{SCR}}$ respectively\;
    }
\end{algorithm}
}
\begin{algorithm}
    \caption{MTSGL Inference}\label{algorithm2}
    \KwData{Input Image $I$}
    % \KwResult{Predict result}
    $f$ $\leftarrow$ Extract feature using encoder $\bm{\theta}$\;
    Predict category using classification head $\bm{\theta}_\mathrm{CLS}$\;
    Predict segmentation using SSA head $\bm{\theta}_\mathrm{SSA}$\;
    % \KwResult{how to write algorithm with \LaTeX2e}
    % initialization\;
\end{algorithm}
\begin{figure*}[!t]
    \centering
    \includegraphics[width=5.2in]{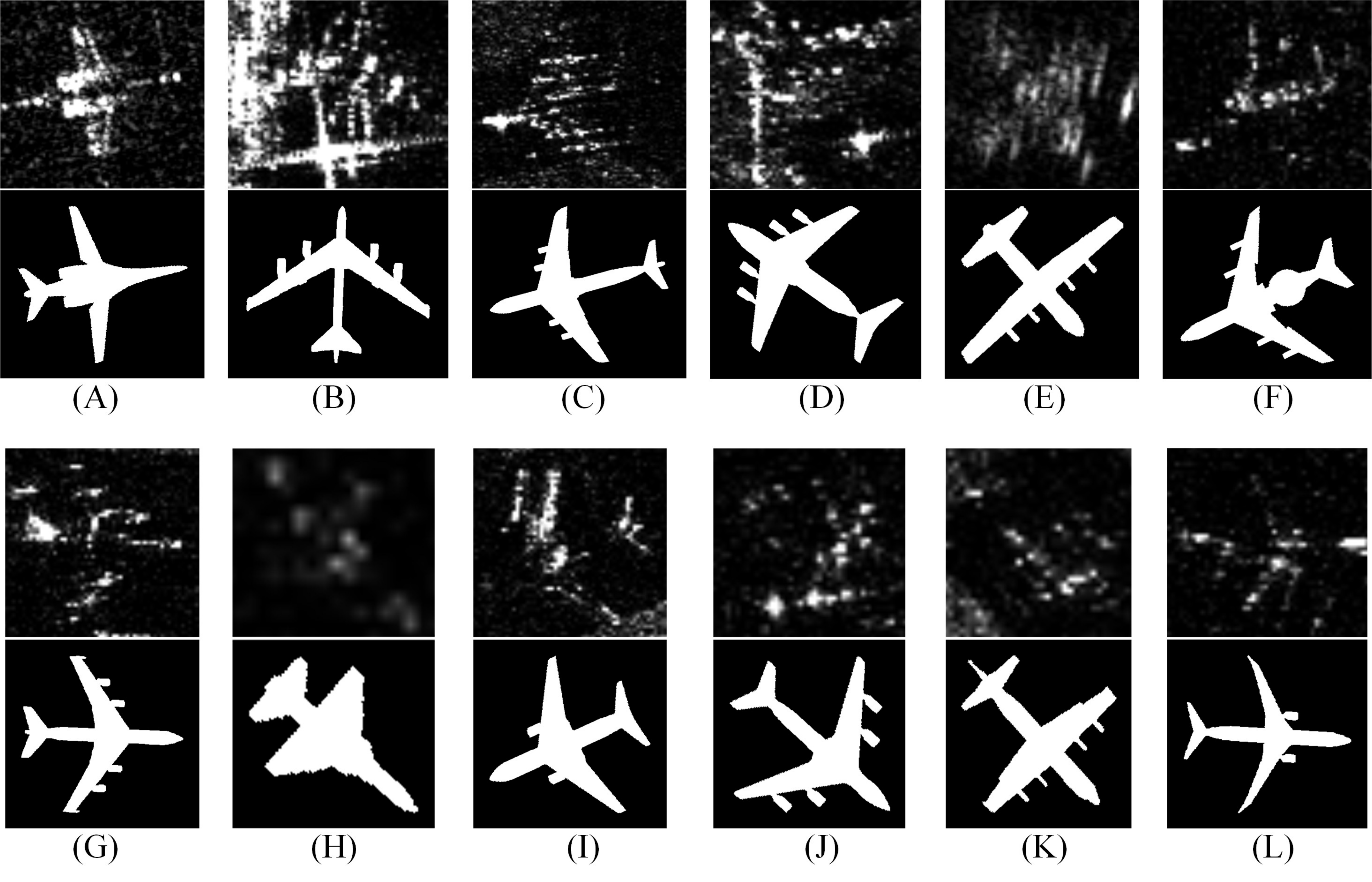}
    \caption{Examples of aircraft and corresponding masks consist of 12 categories. We numbered different types of aircraft with letters from A to L.}
    \label{fig_7}
\end{figure*}

\section{Experiments}
\label{section_IV}
\subsection{Experimental settings}
We conduct extensive experiments on a self-constructed multi-task SAR aircraft recognition dataset (MT-SARD). The MT-SARD dataset contains 68 SAR images, captured from the satellite working in C-band and spotlight mode. All images are orthorectified using the digital elevation model to eliminate the geometric aberrations of oblique imaging. The targets are manually labeled by skilled annotation operators combined with expert interpretation. Consequently, we obtain a total of 2230 aircraft clips, categorised into 12 distinct classes. The binary templates of the aircraft are made according to \cite{RN1580}. Then the aligned masks and the affine projections are manually acquired through the proposed annotation method. Examples and annotations from the dataset are presented in Fig. \ref{fig_7}. \textcolor{black}{The category is carefully labeled according to the corresponding optical images obtained from the Google Earth. The prior knowledge from the optical image could ascertain the presence of the aircraft and its type, thus facilitating the labeling on the SAR image. We also adopt a workflow of multi-person and cross-checking to ensure the reliability of the ground truth.}

\begin{figure}[!t]
    \centering
    \includegraphics[width=3.5in]{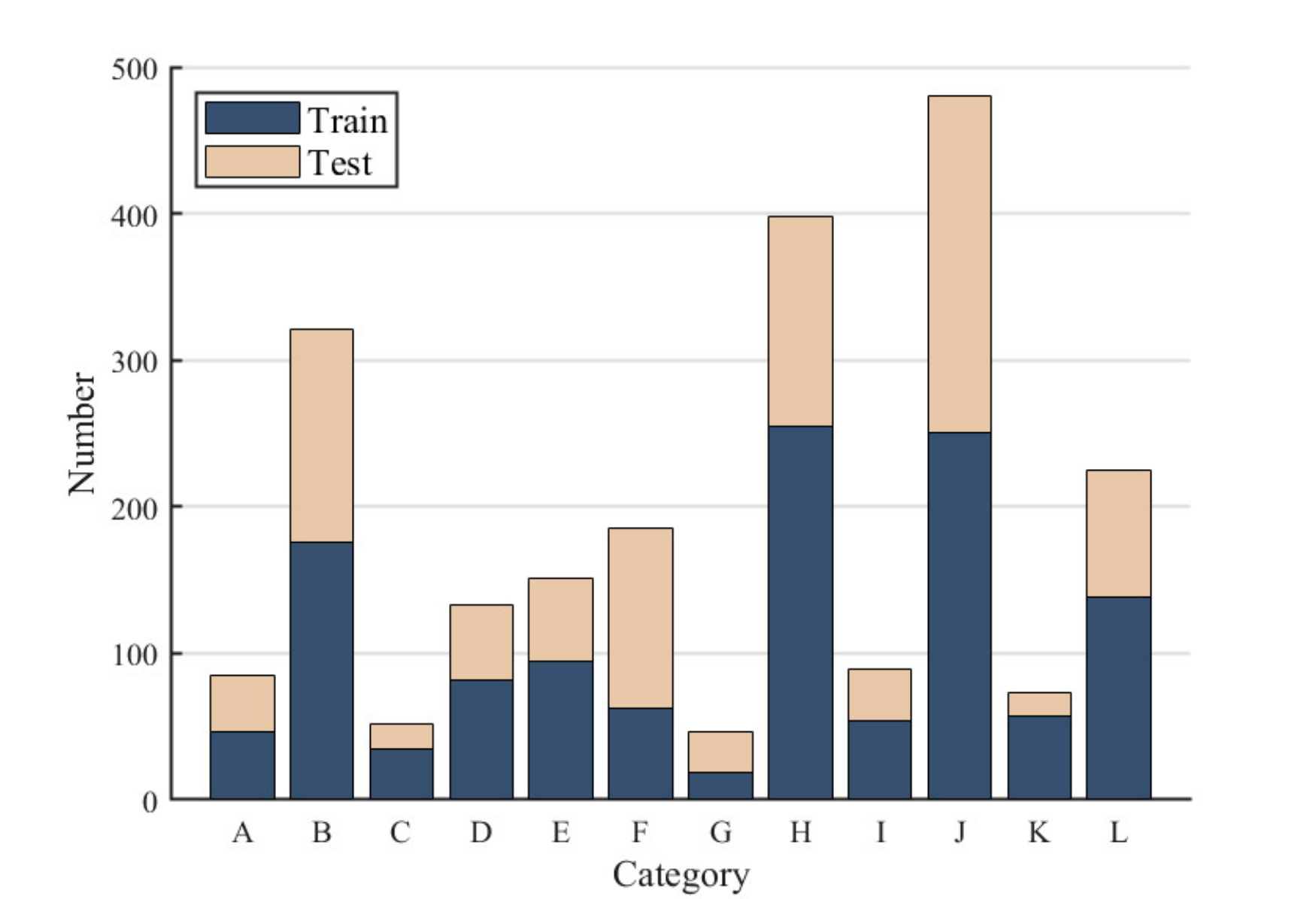}
    \caption{Illustration of the number of each aircraft category from the training and testing data.}
    \label{fig_8}
\end{figure}

\textcolor{black}{Following the 3:2 training-to-testing ratio, total SAR images are divided into 41 for training and 27 for testing. Then we label the aircraft targets using the horizontal bounding box, as shown in Fig. \ref{fig_2}(b). Hence we get 1271 and 968 aircraft clips for training and testing respectively.} The number of different aircraft types is presented in Fig. \ref{fig_8}, which also demonstrates the severe class imbalances.

\textcolor{black}{The Transformer generally exhibits a high demand for training data due to its computational complexity. However, the dataset has fewer samples for effective training compared to the typical CV datasets. Therefore, we adopt the Swin Transformer-Tiny architecture. We further reduce the layer numbers of the Transformer block from [2, 2, 6, 2] to [2, 2, 2, 2]. This results in a total parameter count of approximately 19 million, which is comparable to that of ResNet-34. Furthermore, all the parameters in our Swin Transformer Block are finetuned from the official parameters that have been pretrained under the ImageNet-1k dataset.} The proposed method is trained using the AdamW optimizer with a learning rate of 0.0001 and a weight decay of 0.05. The training process is conducted for 9000 iterations, with a batch size of 32. All experiments are implemented on an 11-GB NVIDIA 2080 GPU server with a PyTorch framework.
  
\subsection{Evaluation Metrics}
In addition to training on the full data, we construct two different training subsets with smaller sizes and leave the testing set unchanged. Two subsets have been established by randomly extracting 40\% and 70\% samples respectively from each class of the whole training dataset. We utilize the overall accuracy (OA) and Kappa coefficient to quantify the performance of the aircraft classification task.
\subsubsection{Overall Accuracy}
OA is typical classification evaluation metric. It measures the ratio of correctly classified targets over all targets as follows:
\begin{equation}
\label{eq15}
\mathrm{OA}=\frac{\sum_{i=1}^C\mathrm{TP}_i}{\sum_{i=1}^C\mathrm{GT}_i}
\end{equation}
where $\mathrm{TP}_i$ and $\mathrm{GT}_i$ are the numbers of correctly classified targets and all targets belonging to the $i$-th category.
\subsubsection{Kappa}
The Kappa coefficient mitigates the imbalance issue by ensuring consistency across categories. It is calculated as 
\begin{equation}
\label{eq16}
\mathrm{Kappa}=\frac{\mathrm{OA}-p_\mathrm{e}}{1-p_\mathrm{e}}
\end{equation}

\begin{equation}
\label{eq17}
p_\mathrm{e}=\frac{\sum_{i=1}^C\left(\mathrm{GT}_i\times\mathrm{PR}_i\right)}{\left(\sum_{i=1}^C\mathrm{GT}_i\right)^2}
\end{equation}
where $\mathrm{PR}_i$ denote the predicted number
of the $i$-th category.

\subsection{Ablation Experiments}

\begin{table*}[!t]
\normalsize
\caption{Influence of each task in the proposed multi-task learning method. Best results are bold.\label{tab_1}}
\centering
\begin{tabular}{ccccccccc}
\toprule
\multirow{2}{*}{} & \multirow{2}{*}{+$\mathcal{L}_\text{SSA}$} & \multirow{2}{*}{+$\mathcal{L}_\text{SCR}$} & \multicolumn{2}{c}{100\%} & \multicolumn{2}{c}{70\%} & \multicolumn{2}{c}{40\%} \\ \cmidrule(lr){4-5} \cmidrule(lr){6-7} \cmidrule(lr){8-9}
&  &  & OA(\%) & Kappa & OA(\%) & Kappa & OA(\%) & Kappa \\ \midrule
& $\times$ & $\times$ & 83.37 & 0.8070 & 80.29 & 0.7709 & 75.50 & 0.7126 \\ \midrule
\multirow{3}{*}{\begin{tabular}[c]{@{}c@{}}w/o\\ Pareto\end{tabular}} & \checkmark & $\times$ & 87.56 & 0.8553 & 85.14 & 0.8272 & 79.79 & 0.7639 \\ [4pt]
& $\times$ & \checkmark & 87.56 & 0.8558 & 85.41 & 0.8301 & 78.95 & 0.7546 \\ [4pt]
& \checkmark & \checkmark & 89.32 & 0.8761 & 87.50 & 0.8556 & 83.00 & 0.8017 \\ \midrule
\multirow{3}{*}{\begin{tabular}[c]{@{}c@{}}w/\\ Pareto\end{tabular}} & \checkmark & $\times$ & 89.09 & 0.8737 & 87.33 & 0.8533 & 81.84 & 0.7882 \\ [4pt] 
& $\times$ & \checkmark & 88.45 & 0.8663 & 86.53 & 0.8436 & 81.86 & 0.7900 \\ [4pt] 
& \checkmark & \checkmark & \textbf{90.21} & \textbf{0.8865} & \textbf{89.57} & \textbf{0.8791} & \textbf{85.99} & \textbf{0.8372} \\ \bottomrule
\end{tabular}
\end{table*}

% \FloatBarrier
\begin{figure*}[!t]
    \centering
    \includegraphics[width=5.0in]{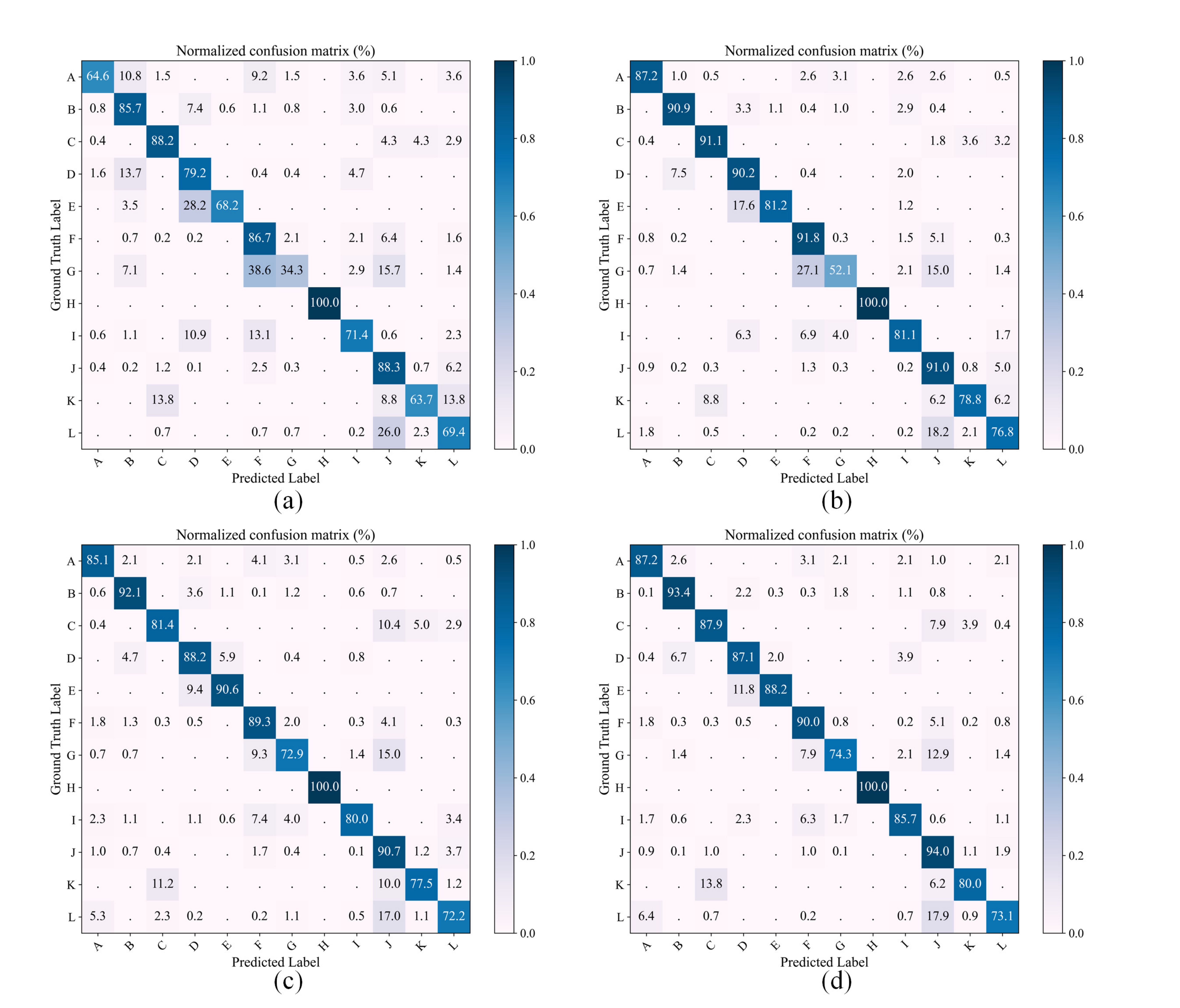}
    \caption{Visualization of the classification results by normalized confusion matrixes. (a) Confusion matrix of the baseline. (b) Confusion matrix of the baseline trained with the SSA module. (c) Confusion matrix of the network trained with the SCR module. (d) Confusion matrix of the network trained with both modules. (b)-(d) are conducted under the Pareto optimization.}
    \label{fig_9}
    \end{figure*}
% \FloatBarrier

In this section, we conduct extensive experiments to evaluate the proposed MTSGL for the SAR aircraft recognition. The ablation experiments are implemented firstly to verify the different tasks' contribution on the baseline network. All studies are performed under the same setting for a fair comparison. 

\textcolor{black}{The overall comparisons are showed in Table \ref{tab_1}. Row 1 indicates the baseline. While rows 2-3 are the w/o Pareto results, rows 5-7 are the w/ Pareto results. For w/o Pareto experiments, the losses of $\mathcal{L}_\text{SSA}$ and $\mathcal{L}_\text{SCR}$ are combined with predefined weights $w_1=3$, $w_2=3$.} It can be seen that both SSA and SCR could bring obvious improvement compared with the baseline trained only by the classification task. Specifically, the SSA and SCR both improve the baseline from 83.37\% to 87.56\%. It is notable that the task-specific heads are discarded during the testing phase. Hence it is believable that these tasks inject additional comprehensive target knowledge into the encoder network, which facilitates the airplane recognition performance of the classification head. The combination of two above tasks could provide further promotion from 83.37\% to 89.32\%. In addition to the upgrading of accuracy, there is also a great improvement in the Kappa coefficient, which illustrates the superiority of the proposed method. Moreover, the Pareto-oriented optimization enables the attainment of the global optimal solution while circumventing the local extremum on a single task. Therefore, the proposed framework under the Pareto optimization finally achieves 90.21\% accuracy, which is a notable increase of 6.84\% in comparison to the baseline. The classification confusion matrices of various ablation studies are shown in Fig. \ref{fig_9}. The recognition accuracy of type H is 100\% due to the considerable disparity in size and structure. For the rest aircraft, the additional tasks could increase performance in each category, which demonstrates the advancement of digesting aircraft information from multi-task learning.

\begin{figure*}[!t]
    \centering
    \includegraphics[width=6.0in]{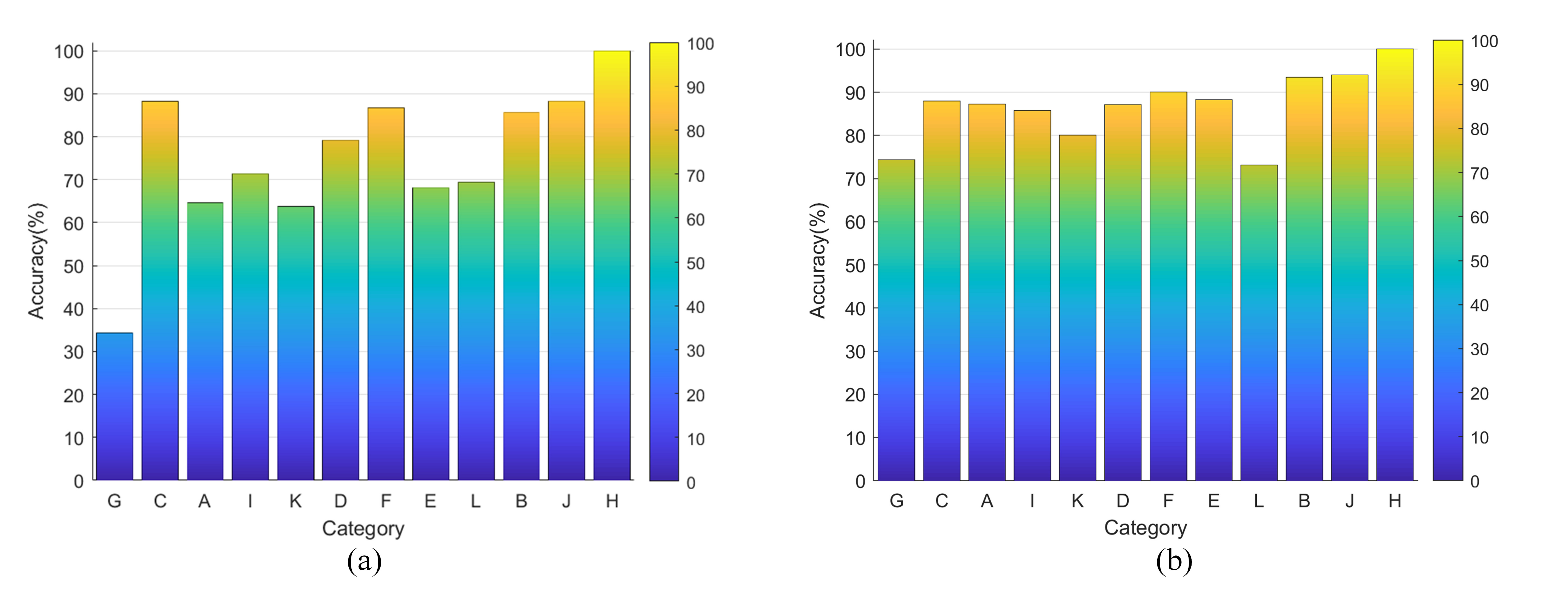}
    \caption{\textcolor{black}{The recognition accuracy for each category, sorted by the number per each category from left to right. (a). Baseline. (b). The proposed MTSGL.}}
    \label{fig_17}
\end{figure*}

\textcolor{black}{The experimental dataset is inherently class-imbalanced. We list the recognition accuracy per each category, as shown in Fig. \ref{fig_17}. The categories are sorted from left to right according to their sample number. It is evident that categories G and H exhibit a marked imbalance issue. The category G has the lowest image number and recognition rate, while the category H is the exact opposite. This imbalanced recognition is mitigated by the proposed multi-task learning method. As demonstrated in Fig. \ref{fig_17}(b), there is a discernible enhancement in recognition for categories with limited samples. This suggests that the aircraft information is effectively exploited in that model could extract enough discriminative features for these categories even with inadequate training samples.}

\textcolor{black}{Besides the class-imbalanced issue, the similar aircraft structural appearance also leads to unsatisfactory recognition. For instance, the class G has the same configuration with the class F and J, e.g., the swept-back wing and four engines. This results in category G being readily misidentified as F and J, as shown Fig. \ref{fig_9}. The proposed SCR is effective in learning robust feature representation, which mitigates the class-imbalanced issue, thus significantly enhancing the recognition of the class G. But subtle structural differences continue to be the cause of the lowest recognition accuracy. This is illustrated in Fig. \ref{fig_17}, which shows that while class G appears to have a higher accuracy compared to the baseline, it still performs lower than other categories. Category L and Category J are very similar in length, width and wing structure, thus resulting in the former being easily recognized as the latter. In this sense, the structural comprehension facilitated by the SSA and SCR is found to be limited for enhancing the recognition for class L.}

We additionally explore the performance under training datasets of various size. In terms of 70\% and 40\% training datasets, the final model achieves 9.28\% and 10.49\% higher than the baseline method. It can be concluded that the proposed multi-task paradigm is more beneficial in cases where training samples are limited. This experimental result also provides a valid indication that enhancing the quality of the annotation, rather than the quantity of the sample, represents an effective alternative in the scenario where a limited training dataset is available for SAR aircraft recognition.

\subsection{Comprehensive Visualization Analysis}\label{Section_IV_D}
To further demonstrate the effectiveness of the proposed framework, we provide a series of visualizations and analysis.

\begin{figure*}[!t]
\centering
\includegraphics[width=6.0in]{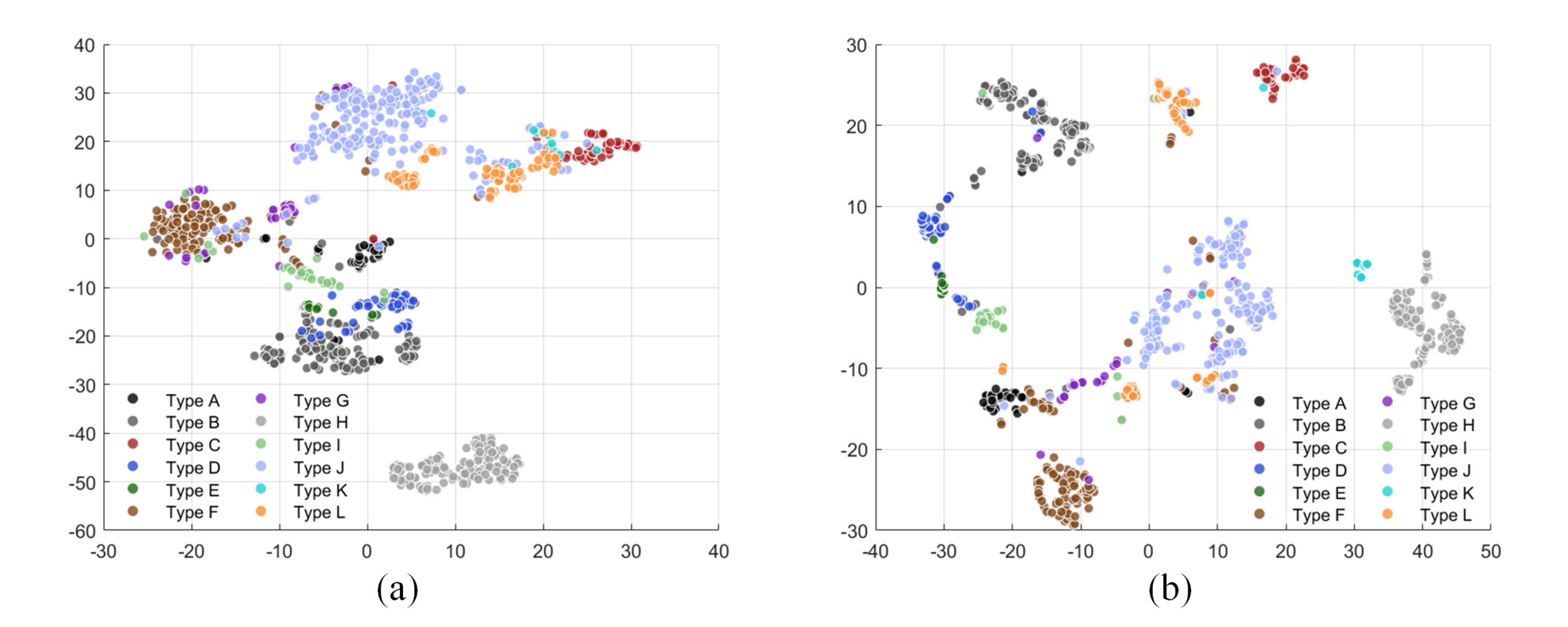}
\caption{Visualization of the feature space using t-SNE for all categories with different colors. (a) Baseline. (b) The proposed MTSGL.}
\label{fig_10}
\end{figure*}
    
% \FloatBarrier
\begin{figure}[!t]
\centering
\includegraphics[width=3.3in]{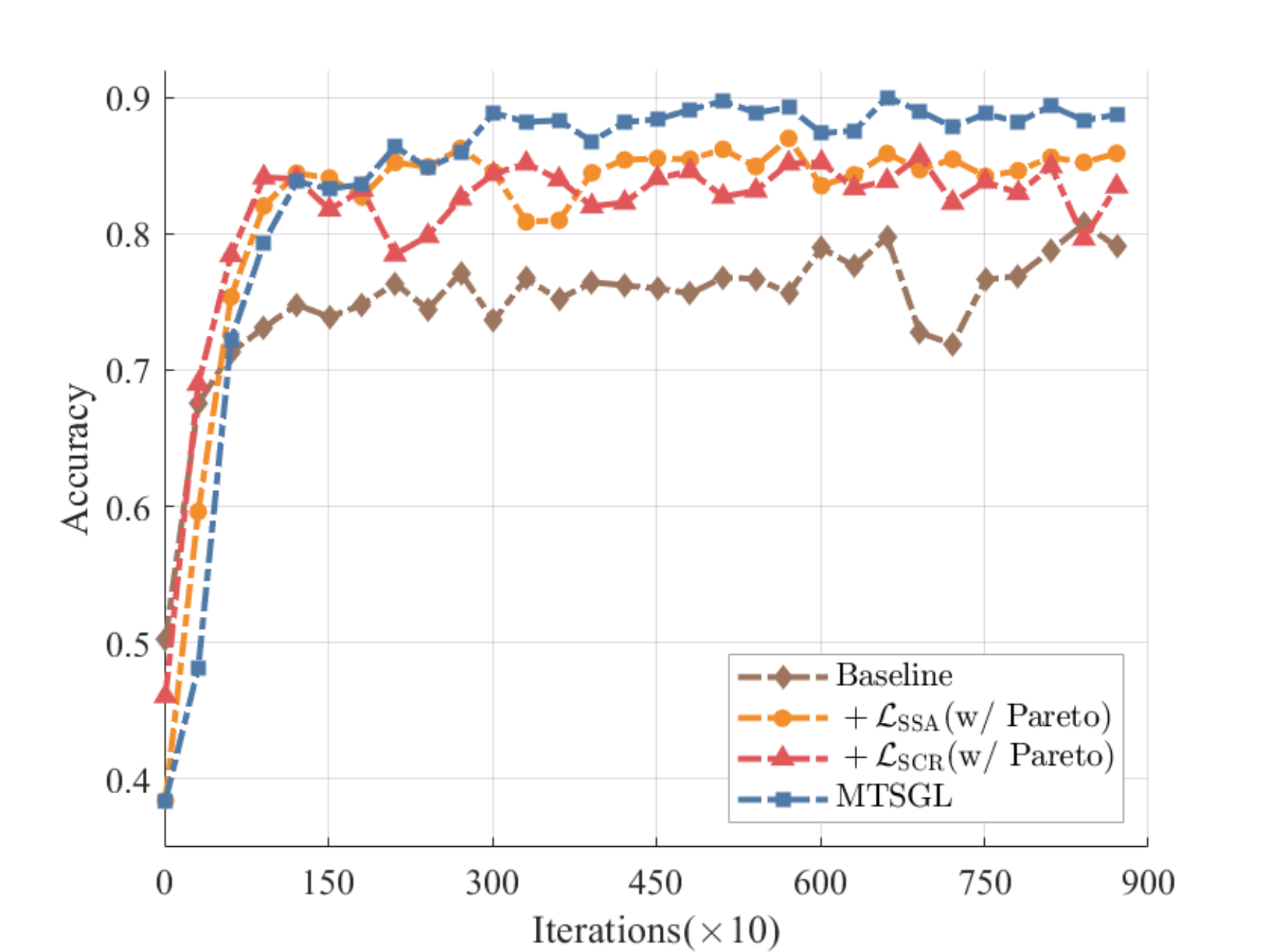}
\caption{Accuracy curves of different task components on the testing dataset.}
\label{fig_11}
\end{figure}
% \FloatBarrier

% \FloatBarrier
\begin{figure}[!t]
\centering
\includegraphics[width=3.3in]{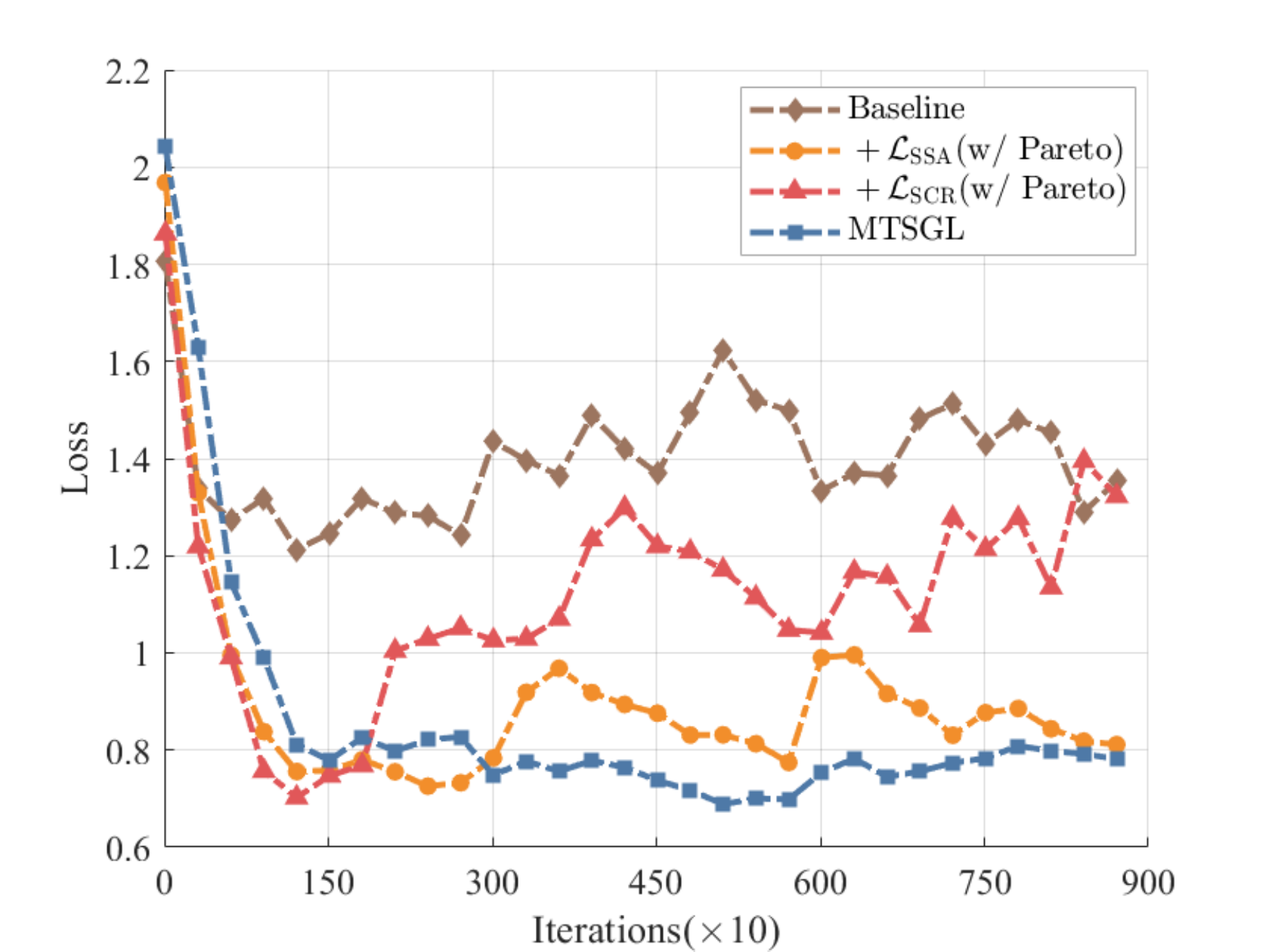}
\caption{Classification loss curves of different task components on the testing dataset.}
\label{fig_12}
\end{figure}
% \FloatBarrier

\begin{figure}[!t]
\centering
\includegraphics[width=3.4in]{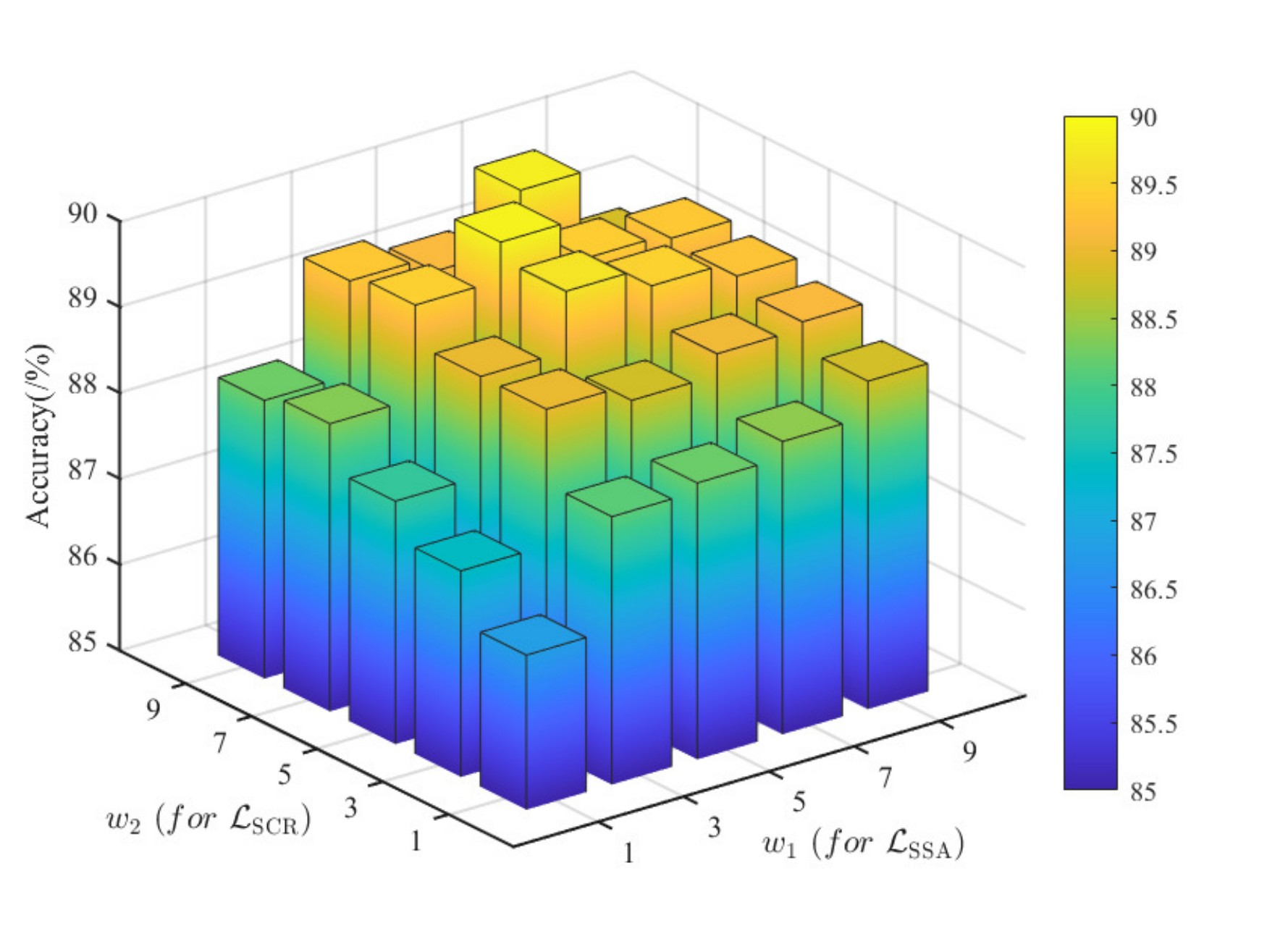}
\caption{Classification accuracy under different weighted combination of the SSA and SCR module.}
\label{fig_13}
\end{figure}

\begin{figure*}[!t]
\centering
\includegraphics[width=5.0in]{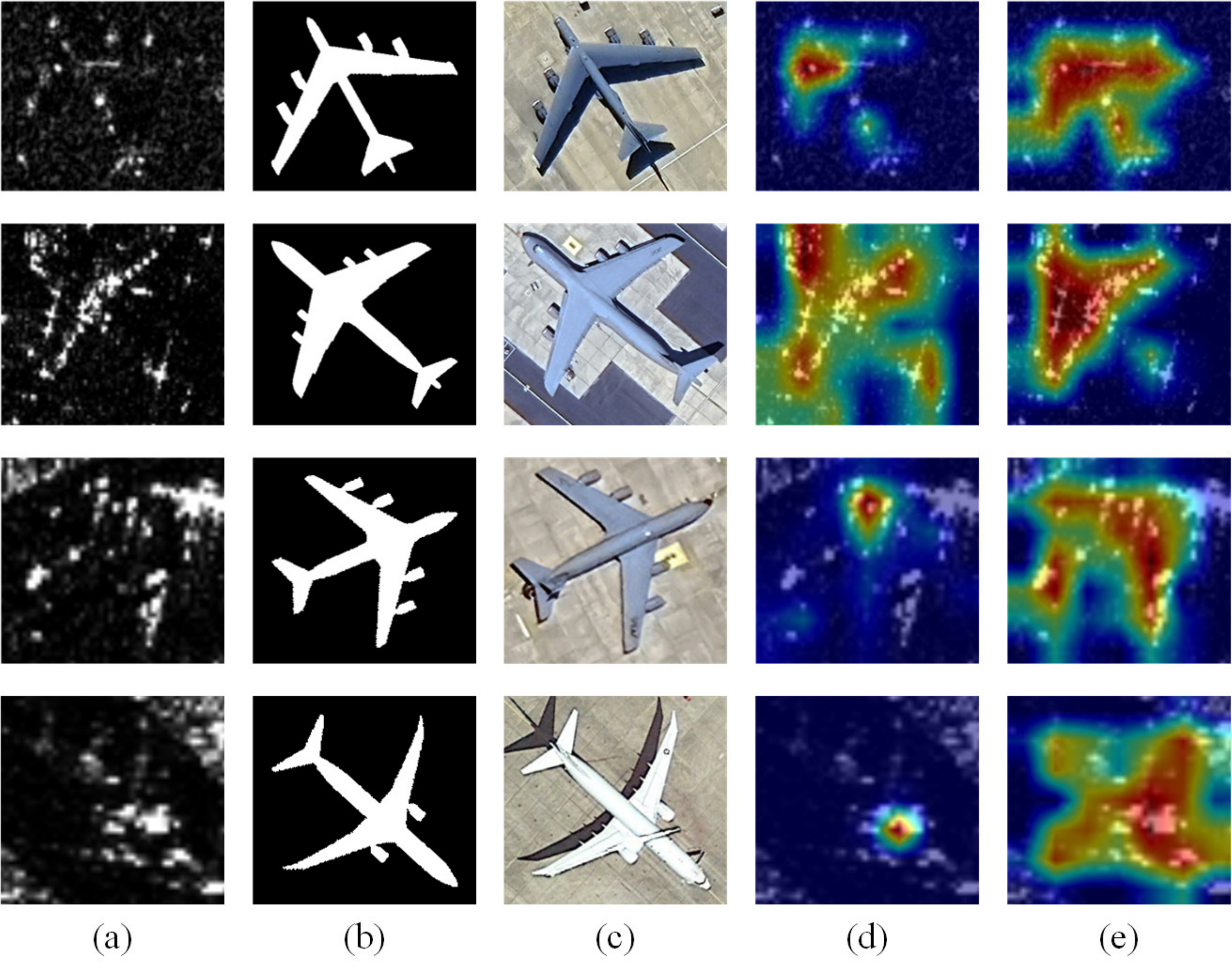}
\caption{Visualization of Score-CAM on testing images of different aircraft categories. From top to bottom are Type B, C, J and L, respectively. (a) SAR images. (b) Optical images. (c) Annotated Masks. (d) Score-CAM from the baseline. (e) Score-CAM from the proposed MTSGL.}
\label{fig_14}
\end{figure*}

% \FloatBarrier
\begin{table*}[!t]
\centering
\normalsize
\caption{Classification and reconstruction results of five different aircraft patches of the MTSGL. \emph{Reconst} means reconstruction result. \emph{Pred} means prediction category.\label{tab_2}}
\centering
\begin{tabular}{m{1.5cm}<{\centering} m{1.5cm}<{\centering} m{1.2cm}<{\centering} m{1.5cm}<{\centering} m{1.2cm}<{\centering} m{1.5cm}<{\centering} m{1.2cm}<{\centering} m{1.5cm}<{\centering} m{1.2cm}<{\centering}}
\toprule
\multirow{2}{*}{Image} & \multirow{2}{*}{Mask} & \multirow{2}{*}{Labels} & \multicolumn{2}{c}{+SSA ($w=3$)} & \multicolumn{2}{c}{+SSA (w/ Pareto)} & \multicolumn{2}{c}{+SSA +SCR (w/ Pareto)} \\ \cmidrule(lr){4-5} \cmidrule(lr){6-7} \cmidrule(lr){8-9} & &
  & \emph{Reconst} & \emph{Pred} & \emph{Reconst} & \emph{Pred} & \emph{Reconst} & \emph{Pred} \\ \midrule
\begin{minipage}[b]{0.23\columnwidth} \centering
    {\includegraphics[width=0.9\textwidth]{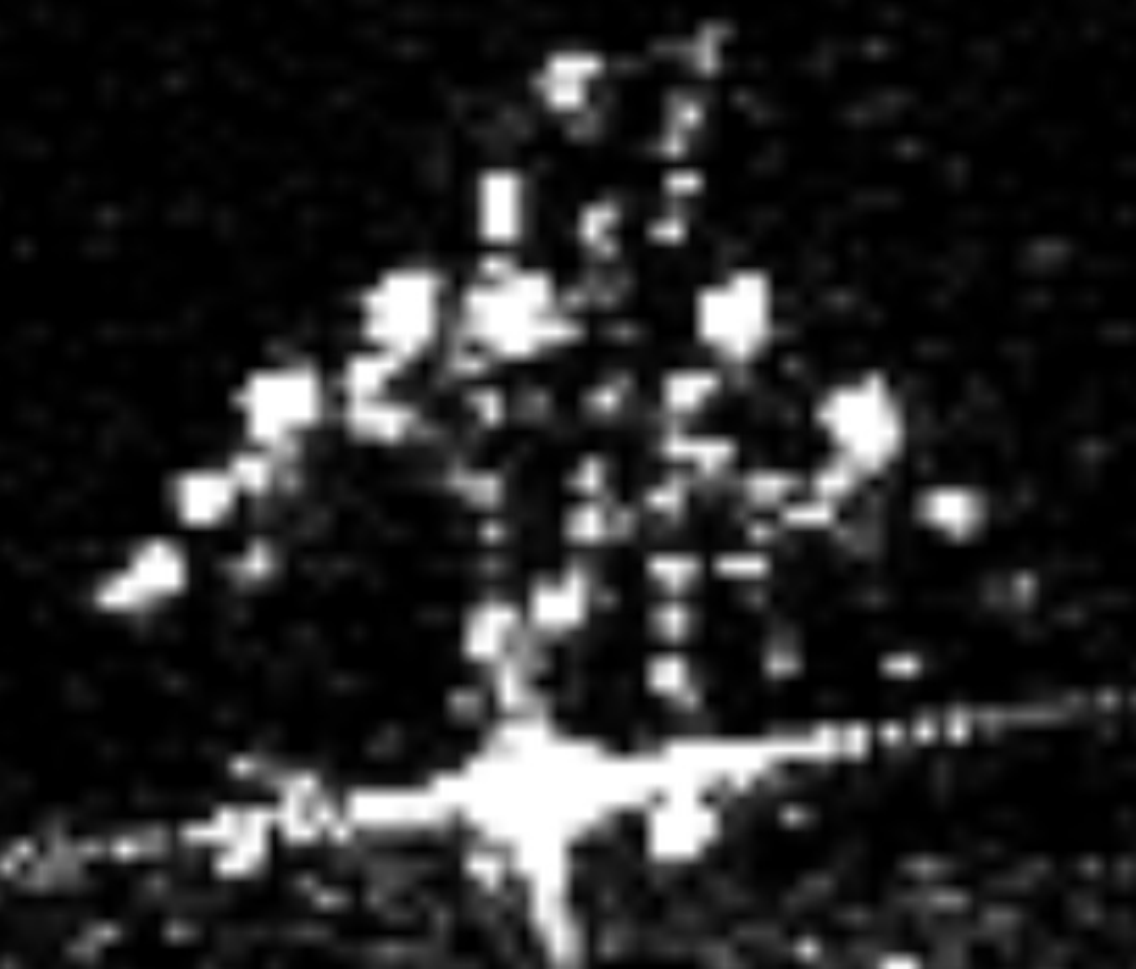}}
\end{minipage}
& \begin{minipage}[b]{0.23\columnwidth} \centering
    {\includegraphics[width=0.9\textwidth]{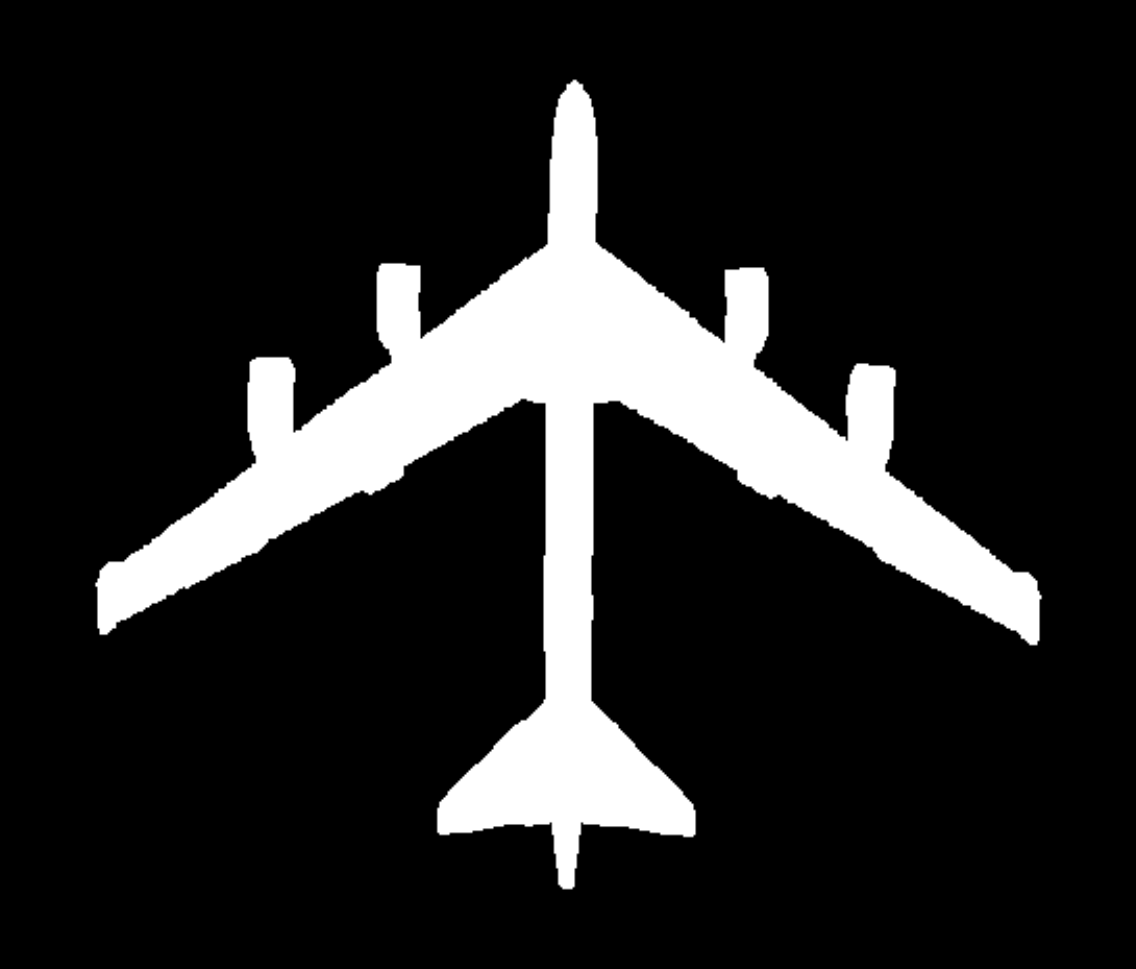}}
\end{minipage} & Type B 
& \begin{minipage}[b]{0.23\columnwidth} \centering
    {\includegraphics[width=0.9\textwidth]{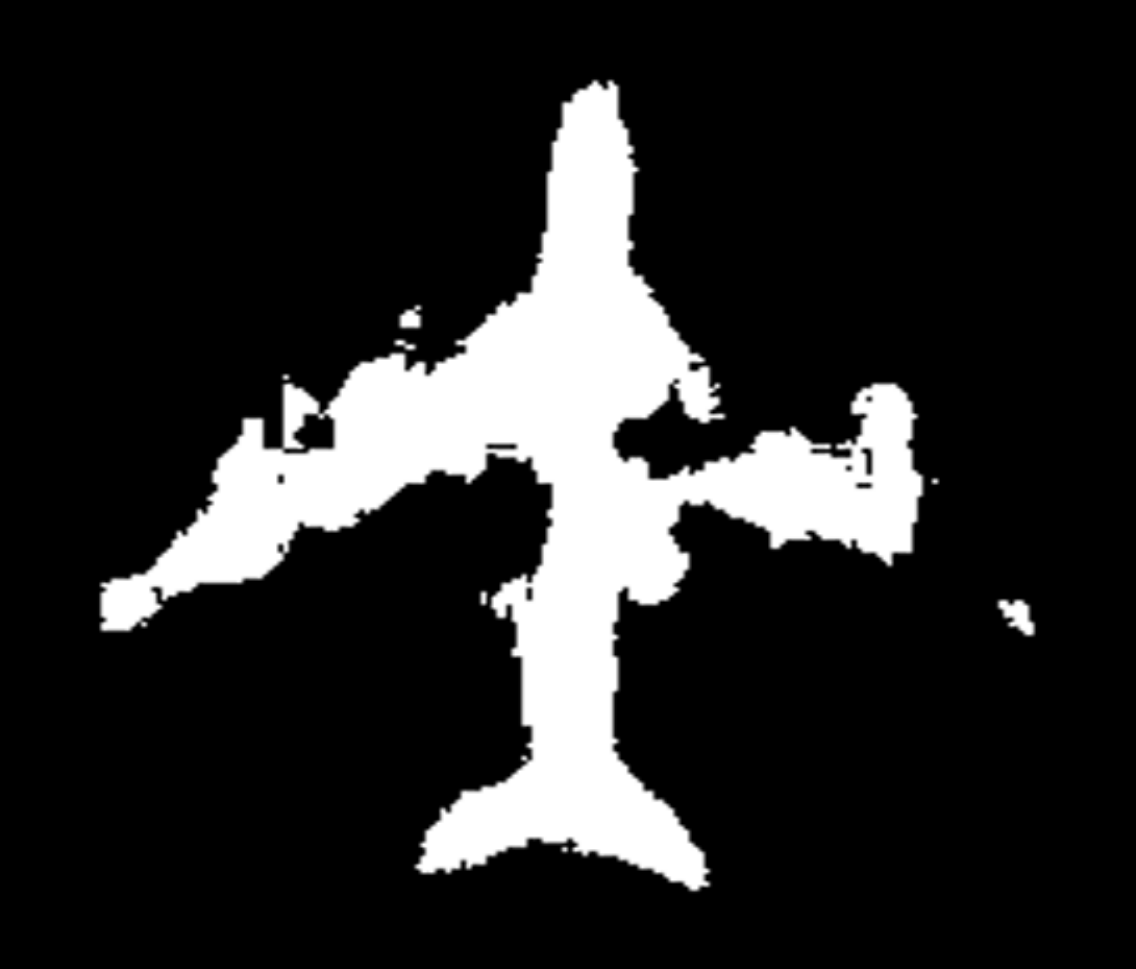}}
\end{minipage} & \textcolor{red}{Type F $\times$} 
& \begin{minipage}[b]{0.23\columnwidth} \centering
    {\includegraphics[width=0.9\textwidth]{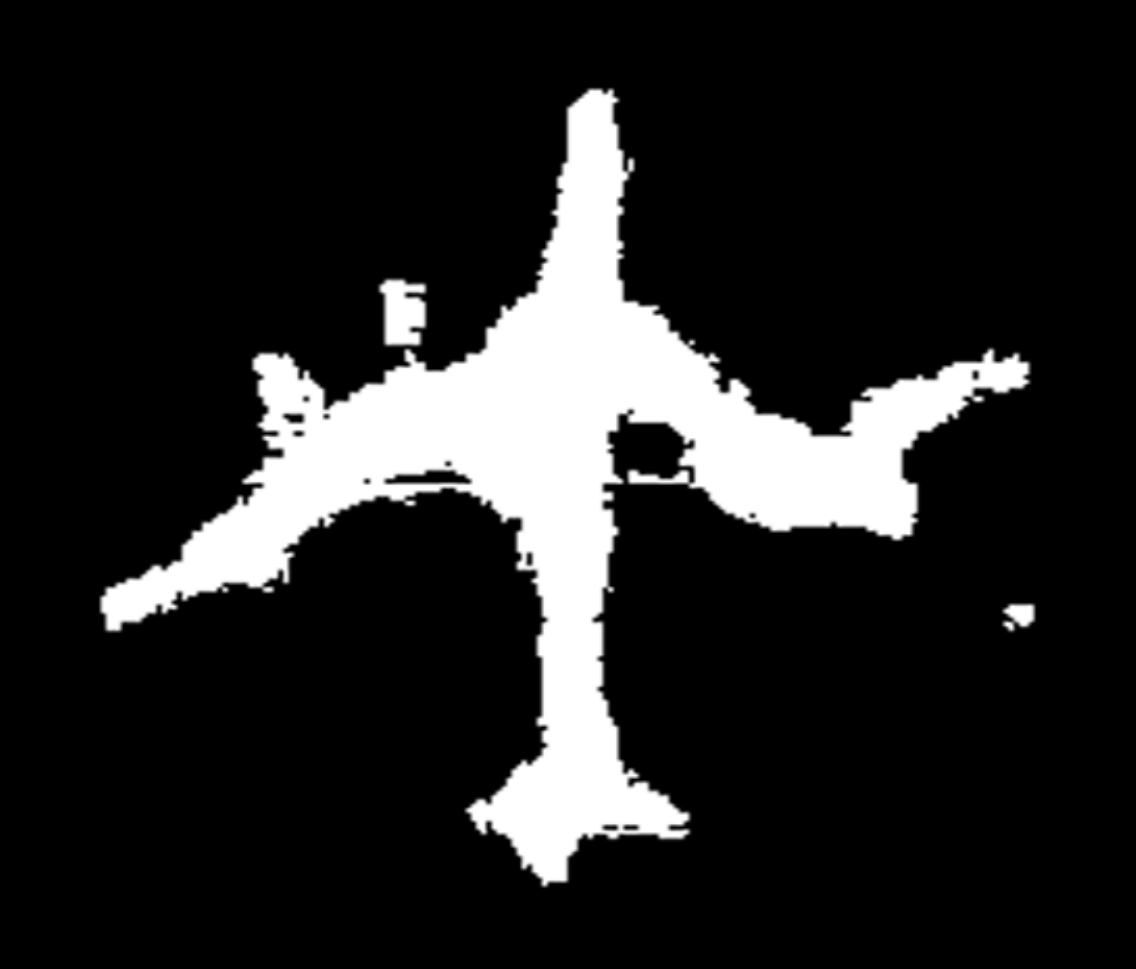}}
\end{minipage} & \textcolor{red}{Type F $\times$}
& \begin{minipage}[b]{0.23\columnwidth} \centering
    {\includegraphics[width=0.9\textwidth]{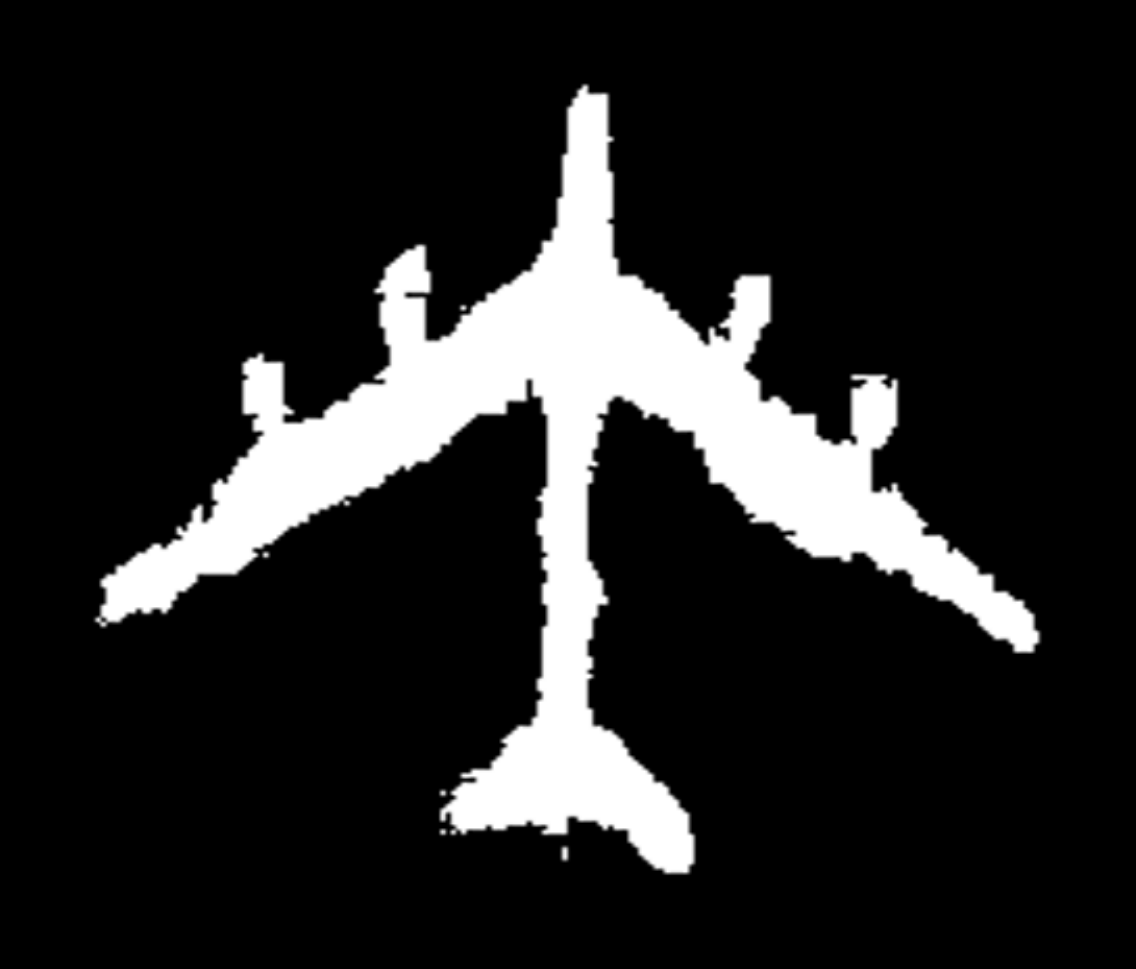}}
\end{minipage} & \textcolor{black}{Type B \checkmark} \\ 
\begin{minipage}[b]{0.23\columnwidth} \centering
    {\includegraphics[width=0.9\textwidth]{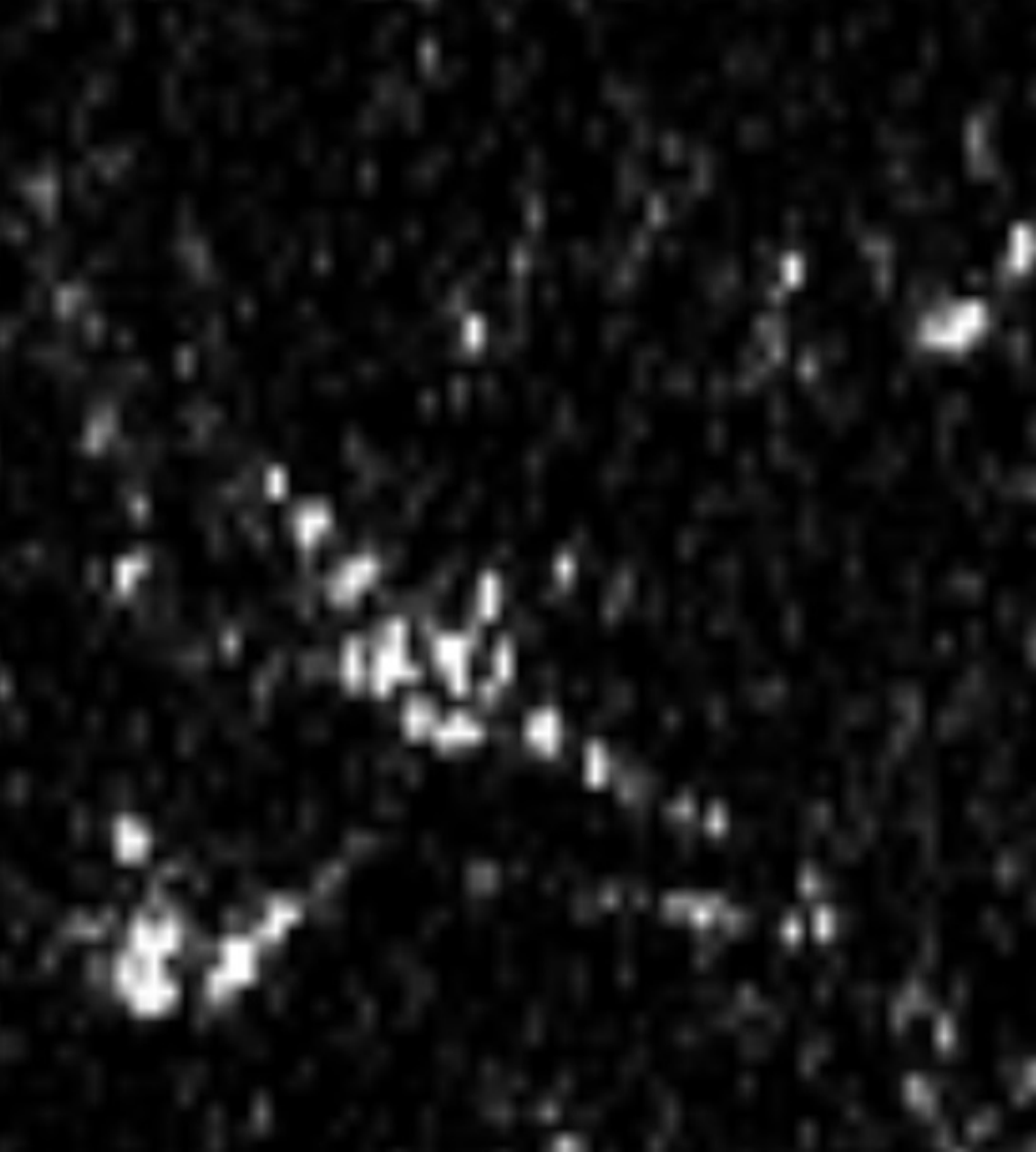}}
\end{minipage} 
& \begin{minipage}[b]{0.23\columnwidth} \centering
    {\includegraphics[width=0.9\textwidth]{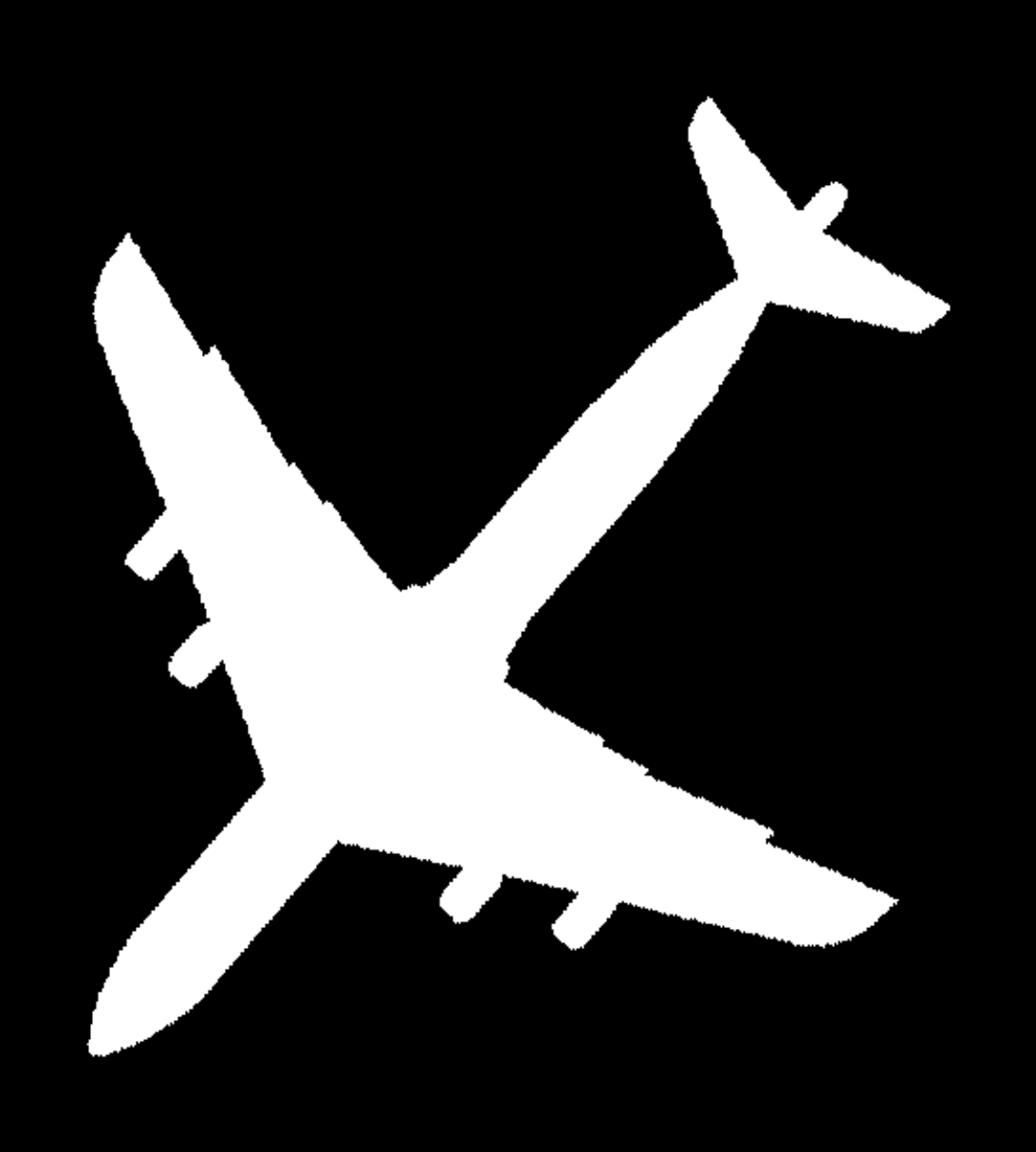}}
\end{minipage} & Type C 
& \begin{minipage}[b]{0.23\columnwidth} \centering
    {\includegraphics[width=0.9\textwidth]{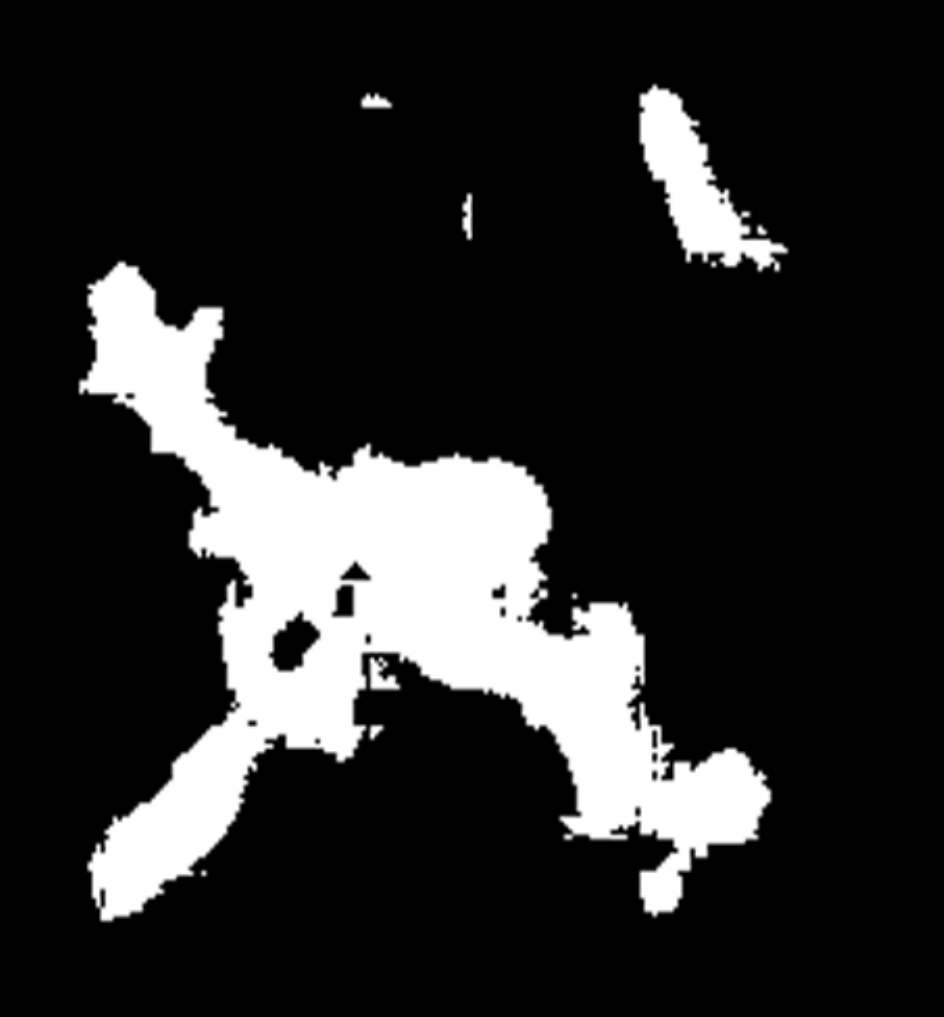}}
\end{minipage} & \textcolor{red}{Type D $\times$}
& \begin{minipage}[b]{0.23\columnwidth} \centering
    {\includegraphics[width=0.9\textwidth]{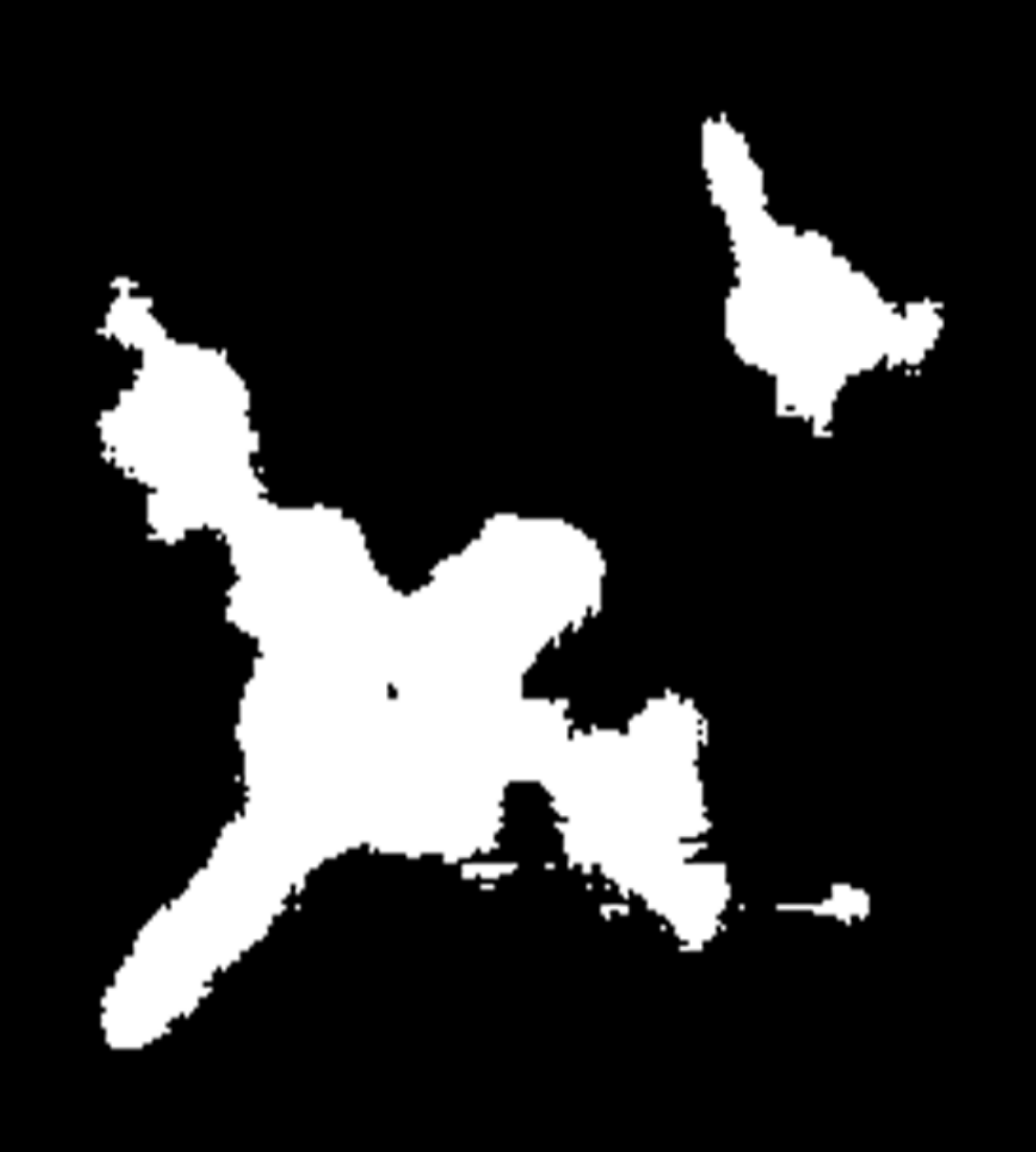}}
\end{minipage} & \textcolor{red}{Type D $\times$}
& \begin{minipage}[b]{0.23\columnwidth} \centering
    {\includegraphics[width=0.9\textwidth]{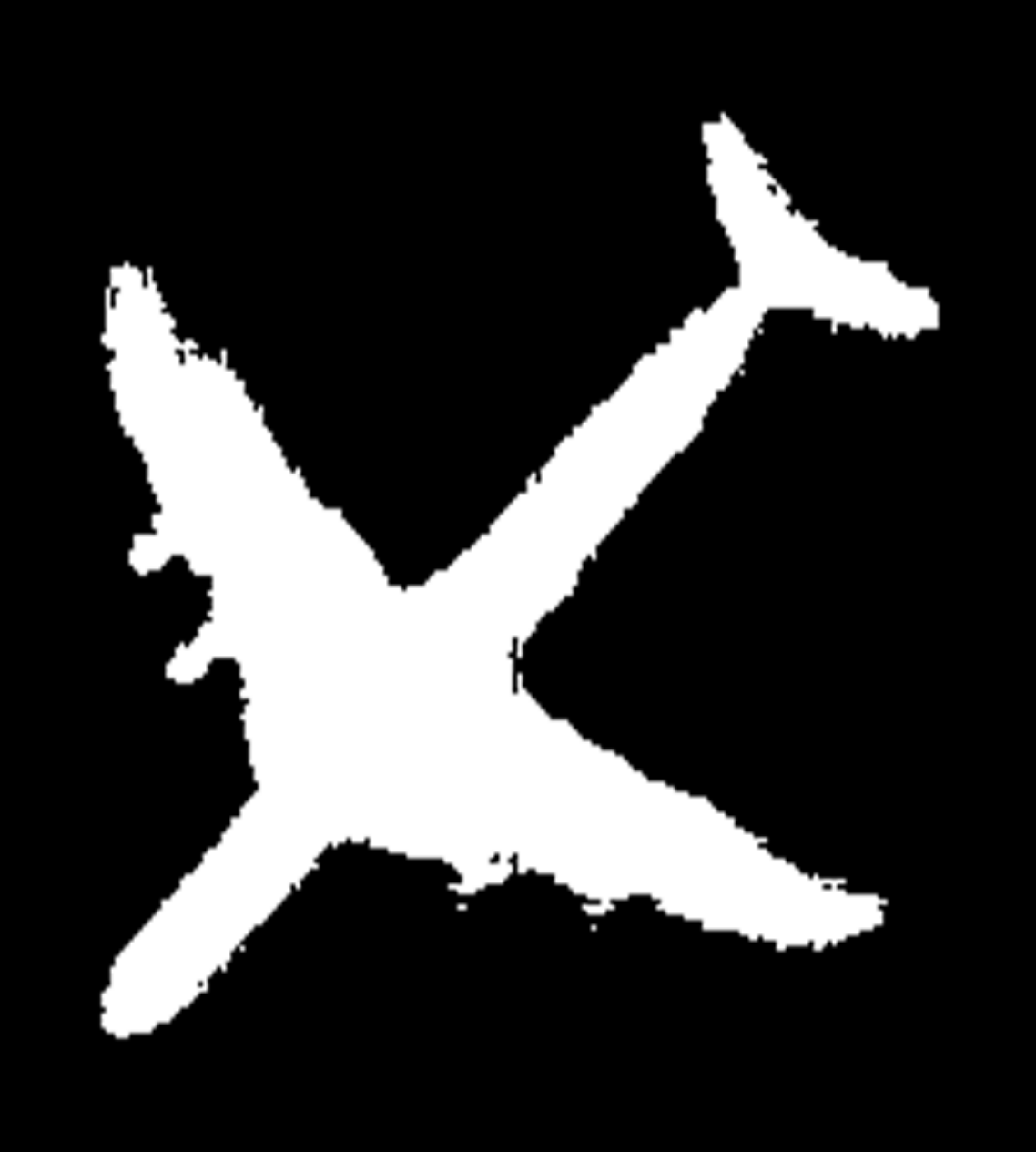}}
\end{minipage} & \textcolor{black}{Type C \checkmark} \\ 
\begin{minipage}[b]{0.23\columnwidth} \centering
    {\includegraphics[width=0.9\textwidth]{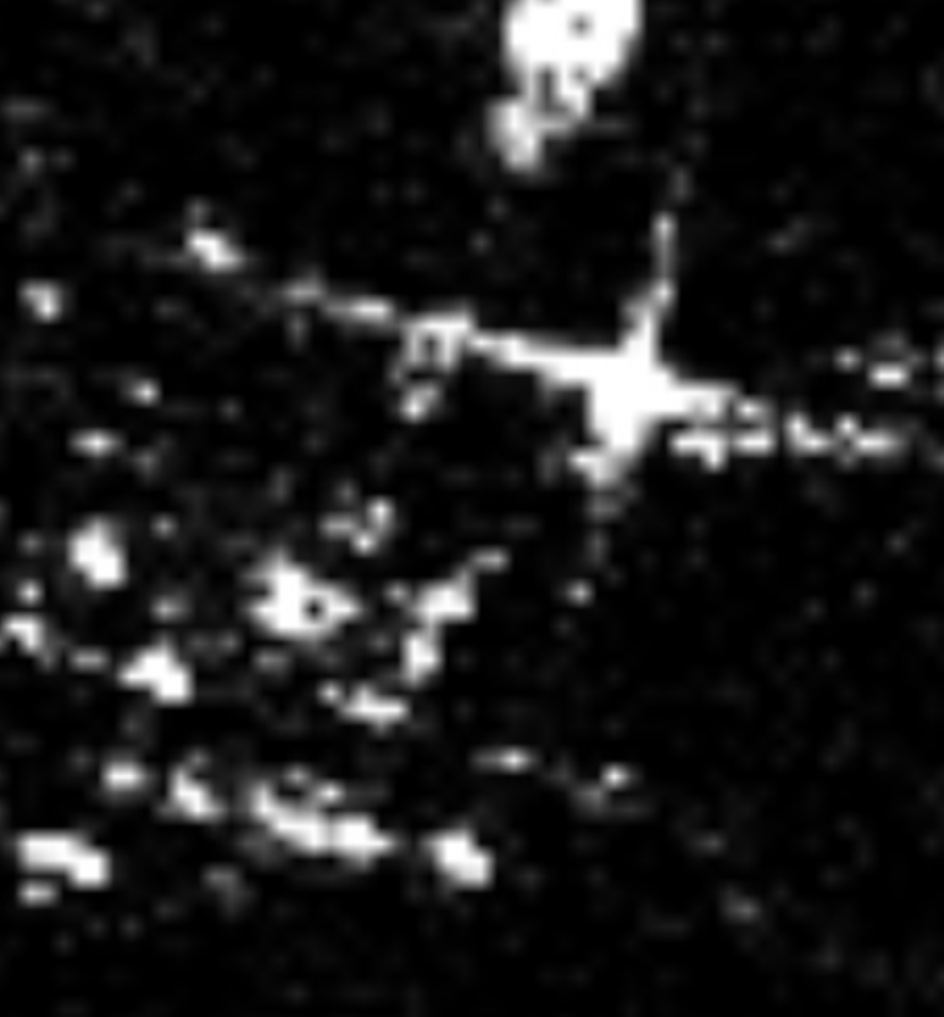}}
\end{minipage} 
& \begin{minipage}[b]{0.23\columnwidth} \centering
    {\includegraphics[width=0.9\textwidth]{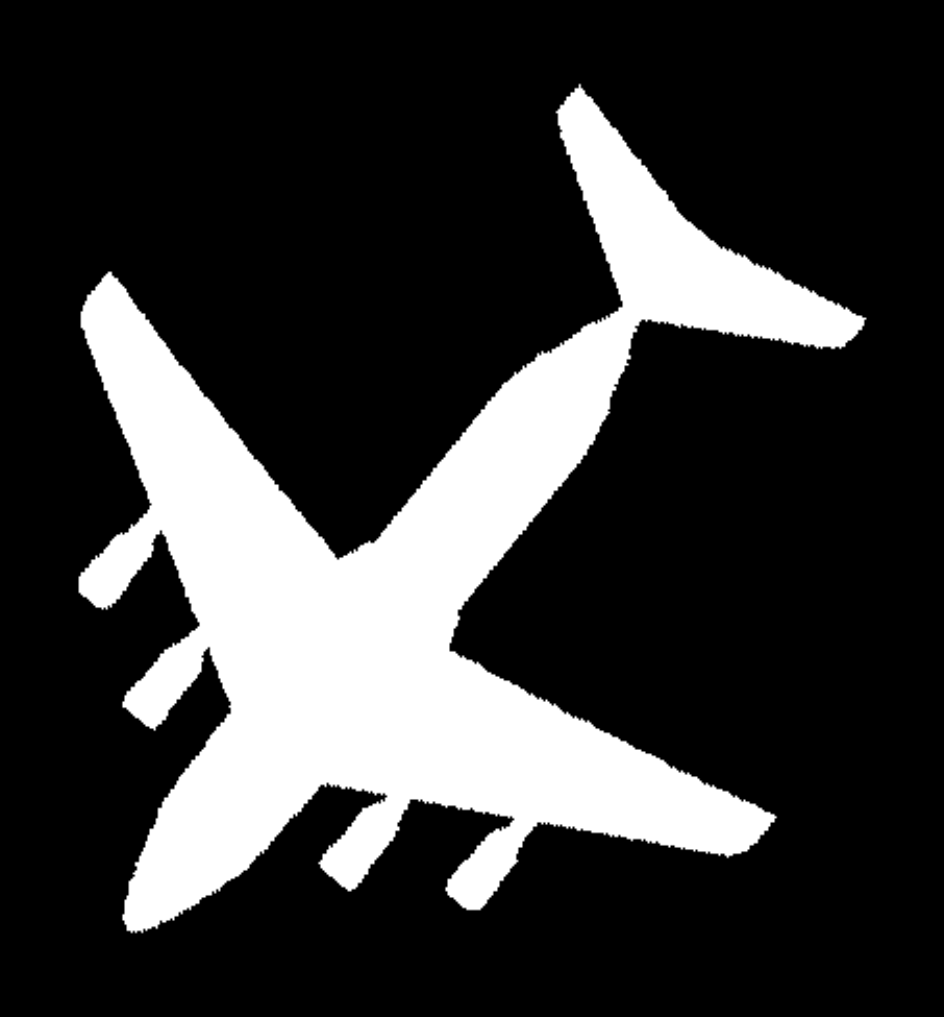}}
\end{minipage} & Type D 
& \begin{minipage}[b]{0.23\columnwidth} \centering
    {\includegraphics[width=0.9\textwidth]{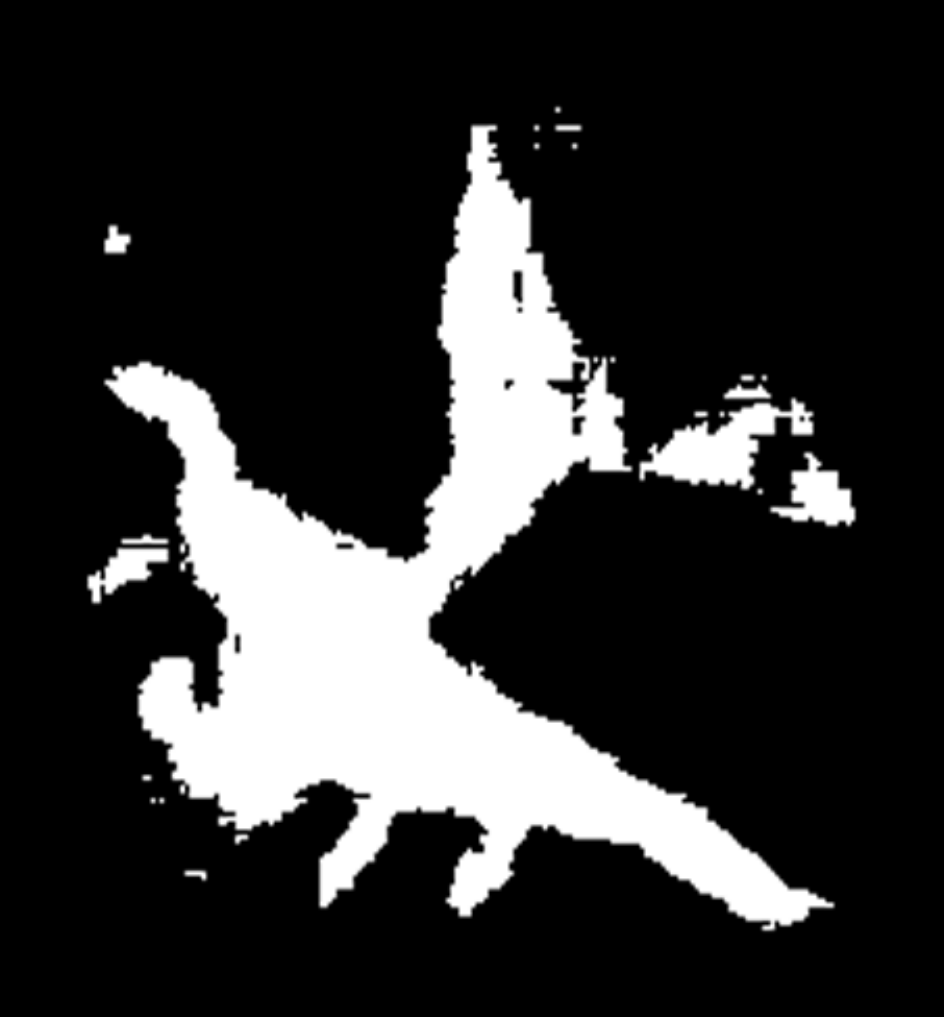}}
\end{minipage} & \textcolor{red}{Type B $\times$}
& \begin{minipage}[b]{0.23\columnwidth} \centering
    {\includegraphics[width=0.9\textwidth]{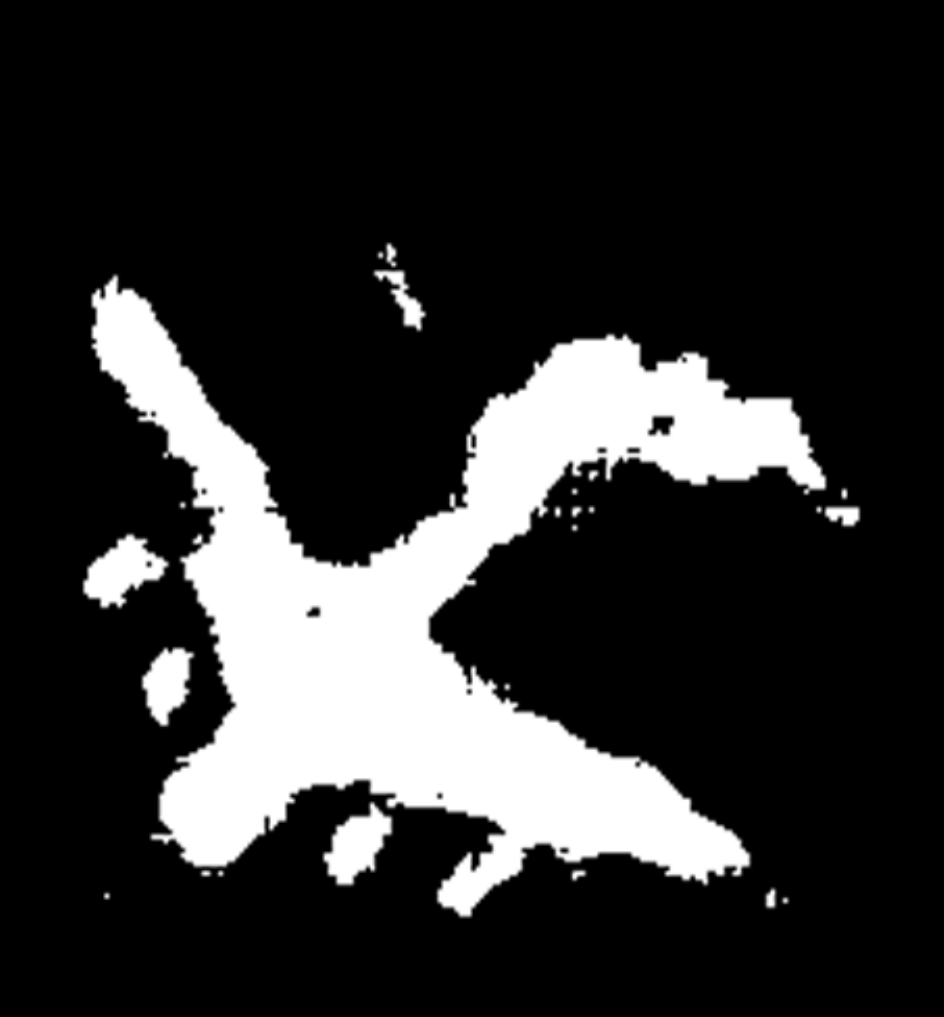}}
\end{minipage} & \textcolor{red}{Type B $\times$}
& \begin{minipage}[b]{0.23\columnwidth} \centering
    {\includegraphics[width=0.9\textwidth]{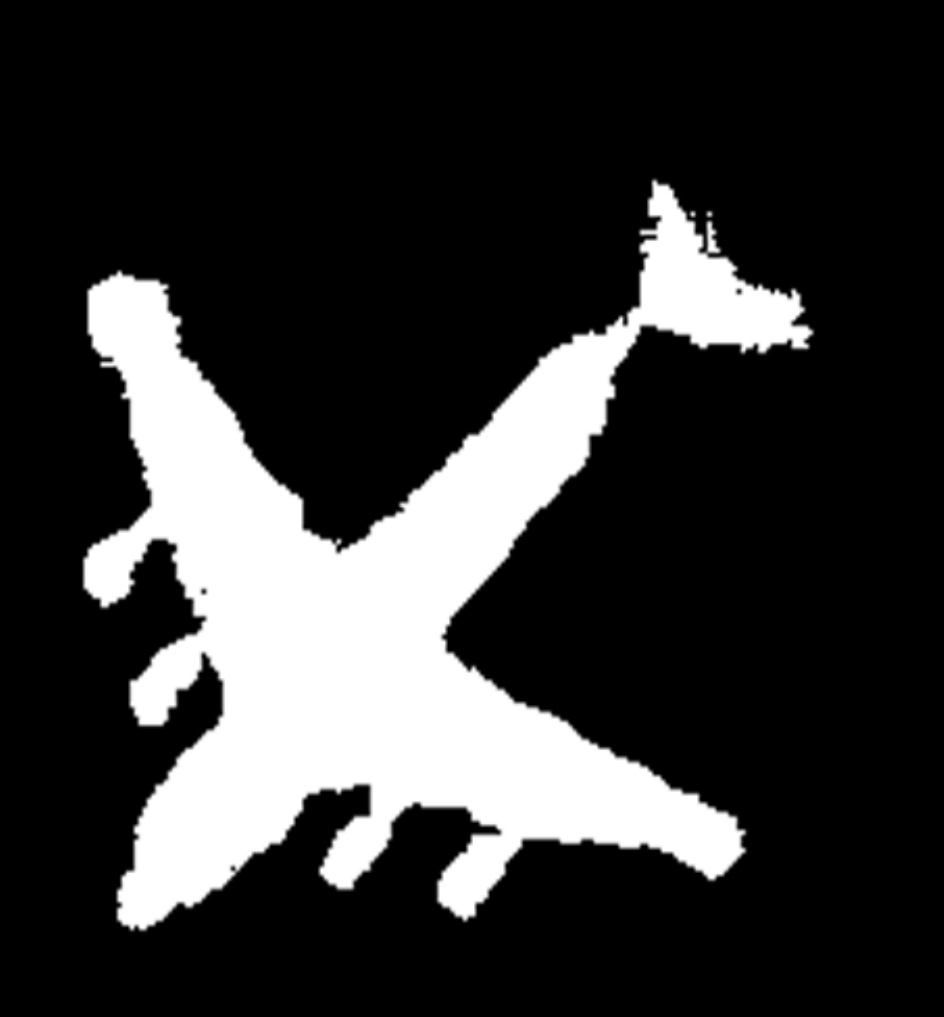}}
\end{minipage} & \textcolor{black}{Type D \checkmark} \\ 
\begin{minipage}[b]{0.23\columnwidth} \centering
    {\includegraphics[width=0.9\textwidth]{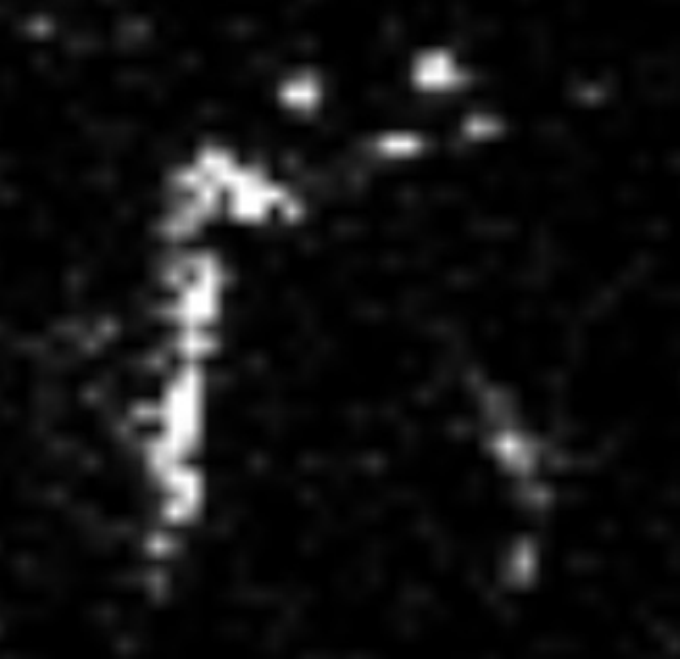}}
\end{minipage} 
& \begin{minipage}[b]{0.23\columnwidth} \centering
    {\includegraphics[width=0.9\textwidth]{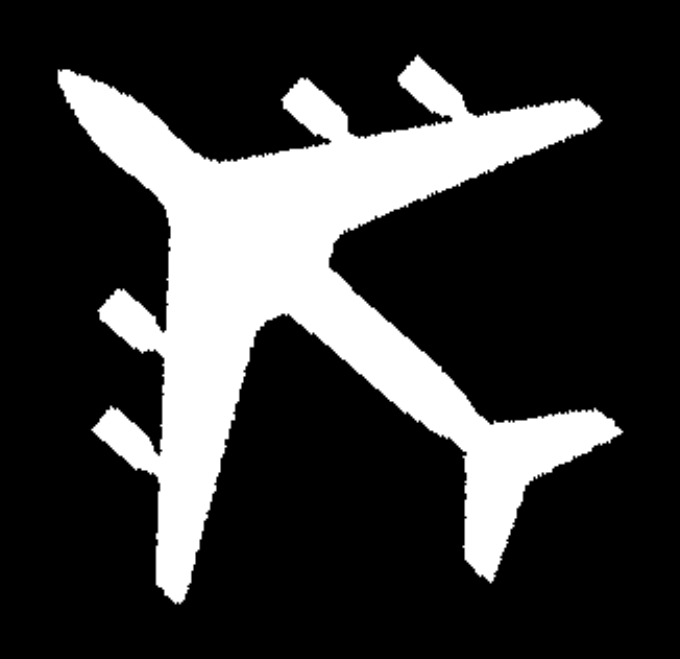}}
\end{minipage} & Type J 
& \begin{minipage}[b]{0.23\columnwidth} \centering
    {\includegraphics[width=0.9\textwidth]{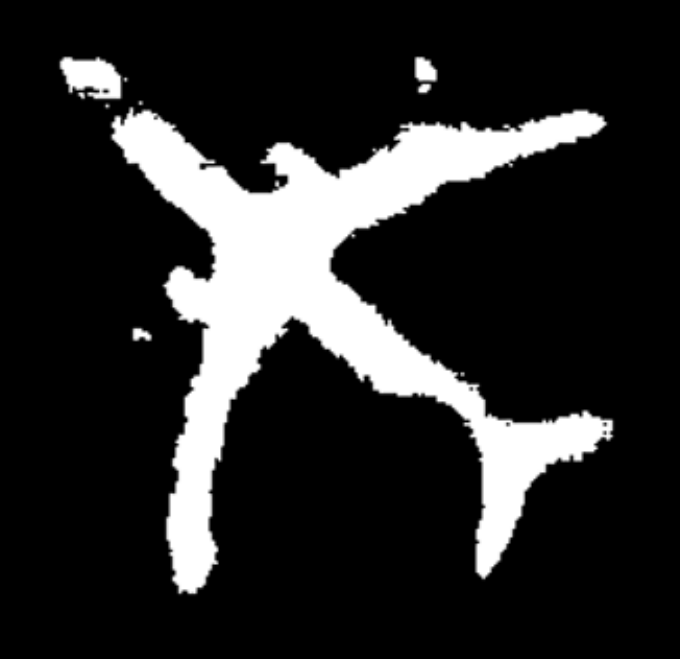}}
\end{minipage} & \textcolor{red}{Type L $\times$} 
& \begin{minipage}[b]{0.23\columnwidth} \centering
    {\includegraphics[width=0.9\textwidth]{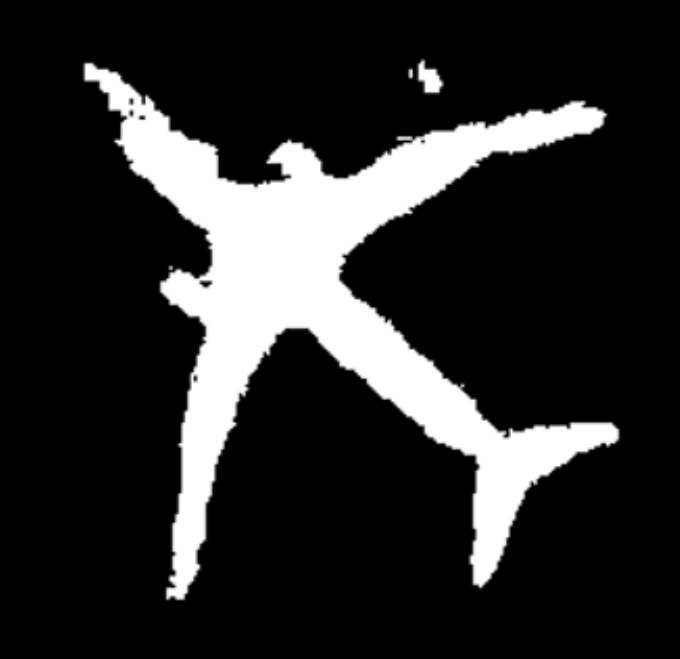}}
\end{minipage} & \textcolor{red}{Type L $\times$} 
& \begin{minipage}[b]{0.23\columnwidth} \centering
    {\includegraphics[width=0.9\textwidth]{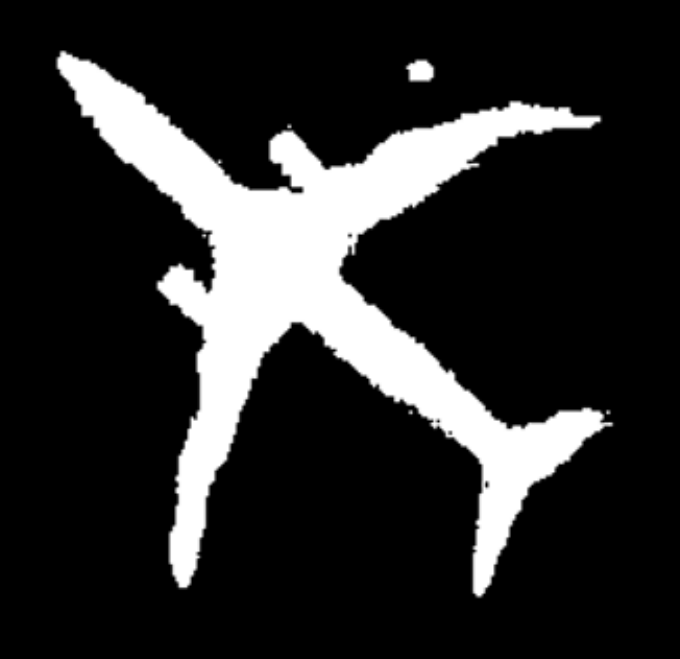}}
\end{minipage} & \textcolor{red}{Type L $\times$} \\ 
\begin{minipage}[b]{0.23\columnwidth} \centering
    {\includegraphics[width=0.9\textwidth]{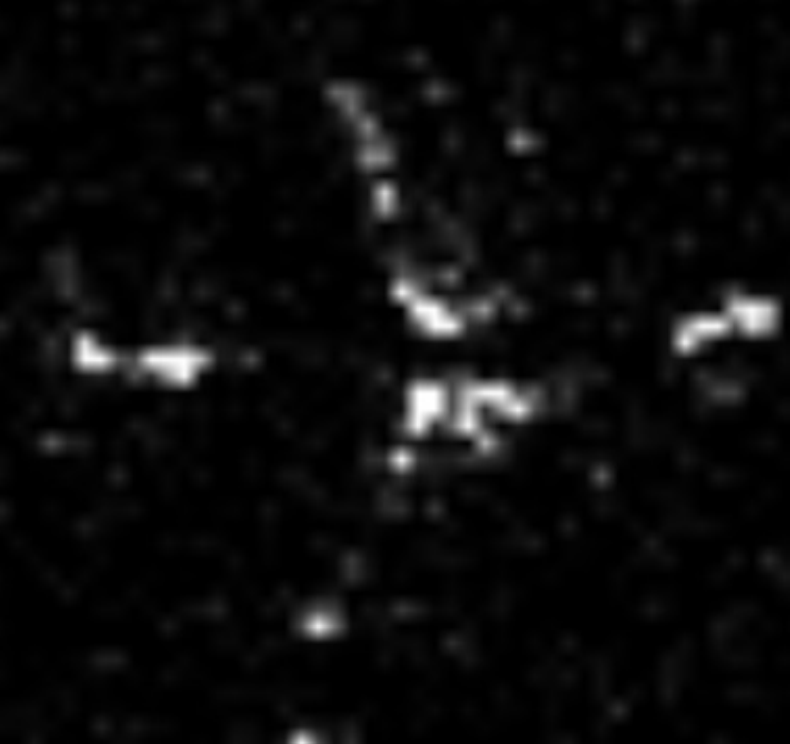}}
\end{minipage} 
& \begin{minipage}[b]{0.23\columnwidth} \centering
    {\includegraphics[width=0.9\textwidth]{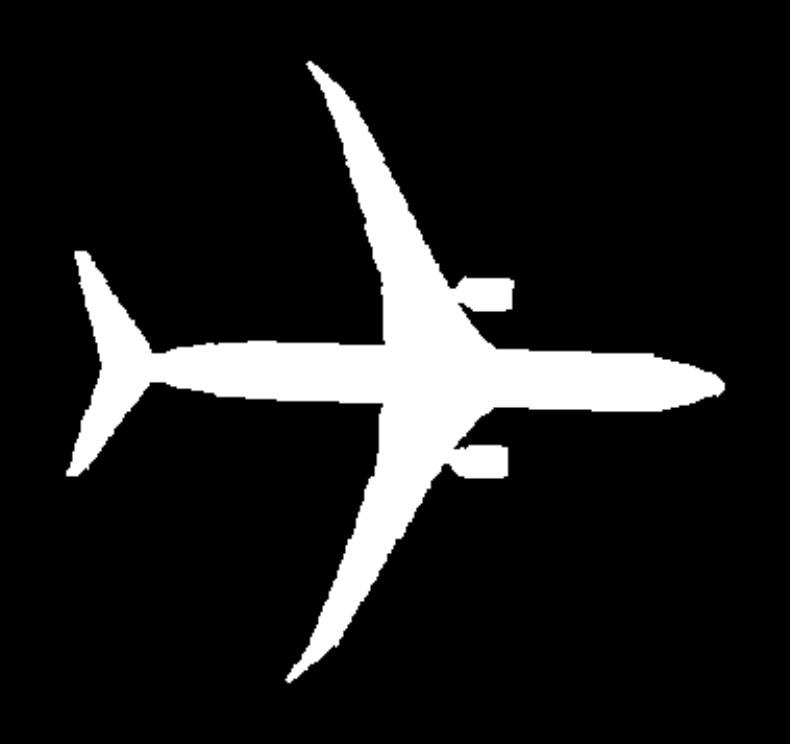}}
\end{minipage} & Type L 
& \begin{minipage}[b]{0.23\columnwidth} \centering
    {\includegraphics[width=0.9\textwidth]{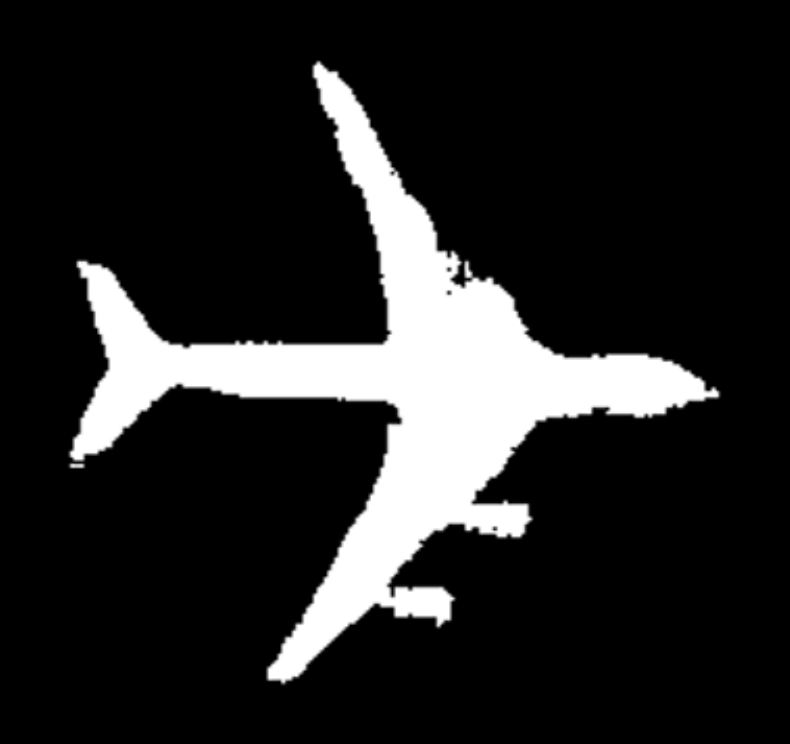}}
\end{minipage} & \textcolor{red}{Type J $\times$}
& \begin{minipage}[b]{0.23\columnwidth} \centering
    {\includegraphics[width=0.9\textwidth]{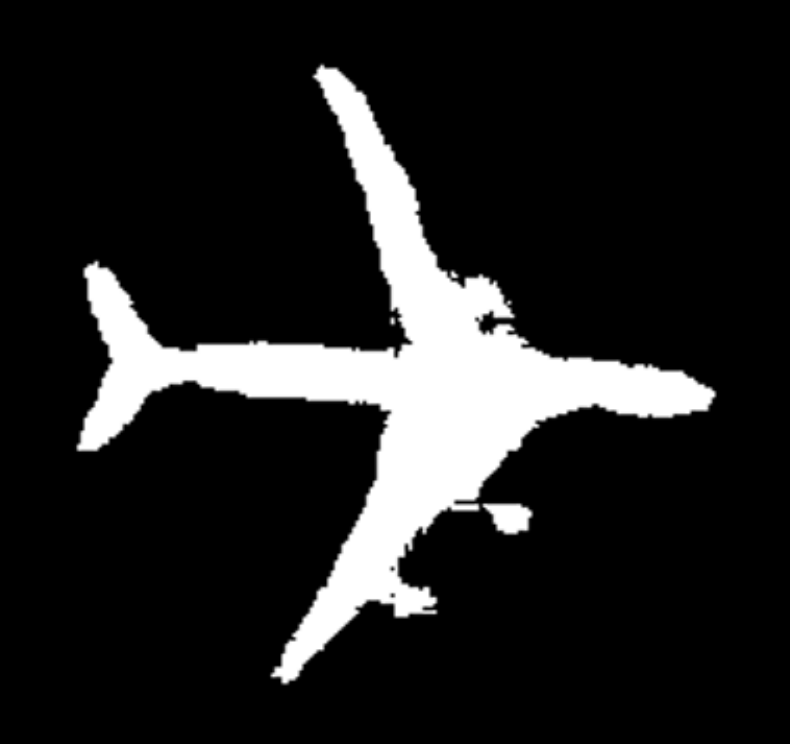}}
\end{minipage} & \textcolor{red}{Type J $\times$}
& \begin{minipage}[b]{0.23\columnwidth} \centering
    {\includegraphics[width=0.9\textwidth]{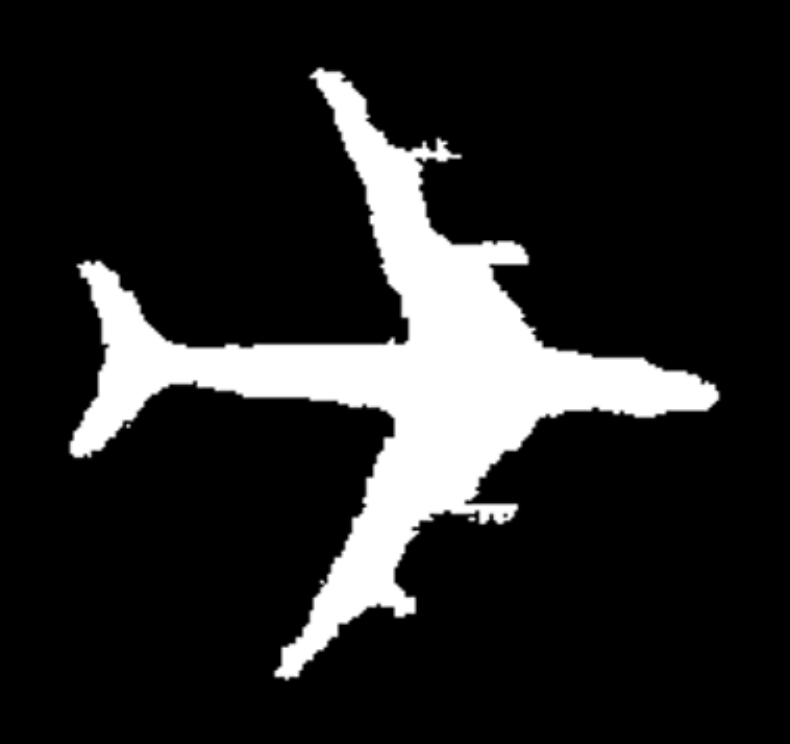}}
\end{minipage} & \textcolor{red}{Type J $\times$} \\ \bottomrule
\end{tabular}
\end{table*}
% \FloatBarrier

The feature visualization using t-distributed stochastic neighbor embedding (t-SNE) \cite{RN1308} is firstly shown in Fig. \ref{fig_10}. \textcolor{black}{The features used to generate the t-SNE plots are derived from the output of the encoder. In the t-SNE visualization, the above conclusion could be intuitively understood. The features of categories G and F are distributed much closer together. The same goes for L and J. Overall, the features from the proposed MTSGL are more aggregated in the same class and separated among different classes. Therefore, the proposed method yields a more discriminative decision with tighter intra-class compactness.}

The curves of both the classification loss and the accuracy on the testing dataset are then displayed in Fig. \ref{fig_11} and Fig. \ref{fig_12}. \textcolor{black}{It can be observed that loss curves for the baseline, $\mathcal{L}_\text{SSA}$ and $\mathcal{L}_\text{SCR}$ exhibit unstable fluctuations at the end training stage. This can be attributed to the overfitting to the limited training datasets, which prevents reaching the optimal minima. Despite that, the SSA and SCR stills provide better optimization compared to the baseline. We argue that if the model is learned in a cognitive process that is consistent with the aircraft knowledge, it will converge more to the better minima. The network trained under the proposed MTSGL guarantees a more stable and smooth convergence to the optimal point.} In consequence, the multi-task consisting of SSA and SCR in principle furnishes a meaningful regularization role when learning a discriminative representation.

It has been shown that the proposed multi-task learning framework could improve the quality of the learned representation and therefore benefit the classification task. While the proposed framework demonstrates an adaptive global convergence, further investigation is warranted to ascertain the precise relationship between classification performance and these tasks. We turn back to the optimization method via averaged loss of weighted multi objective (see Equation \eqref{eq9}). The classification loss weight $w_1$ is set to 1 and the SSA and the SCR weights are set dynamically from 1 to 9. The classification performance influenced by various combination among tasks is demonstrated in Fig. \ref{fig_13}. As a result, the classification accuracy occurs a positively proportional to the weight coefficients assigned to both tasks. This further verifies the effectiveness of excavating the target information from multi-tasks. 

The decision-making rationality is examined using a post-hoc explanation method, called Score-CAM \cite{RN1607}. The attention activation map is derived from the contribution of their highlighted input features to the model output. We choose the last Swin transformer block's output for interpreting the decision. Fig. \ref{fig_14} demonstrates the Score-CAM results derived from two swept-back aircraft and two flat-winged aircraft. As shown in Fig. \ref{fig_14} (d) and (e), different regions have different corresponding attention masks. The hotter the color, the larger the attention value. The highlighted attention of the baseline appears to be weakly related to the target regions. This can be attributed to the fact that the classification task attends only to learning the discriminative boundaries while neglecting the structural attributes inherent to aircraft targets. The proposed method places greater emphasis on the more desirable target regions, demonstrating that the evidence pertaining to the target's structure is responsive to inferential predictions. What's more, the wing and engine areas exhibit the highest attention value across all four aircraft categories. The results demonstrate that the proposed method is capable of understanding the aircraft targets and extracting intuitive and readily comprehensible discrimination features.

The mask decoder can be employed as a feature visualization tool to intuitively ascertain the number and the type of structural components preserved in the encoded features. We recover the mask outputs from the decoder of the networks trained with the fixed-weighted SSA module, the Pareto-oriented SSA module, and the proposed method respectively. The pixels predicted to be non-background are set to 1 and vice versa to 0. As shown in the first three samples in Table \ref{tab_2}, the proposed algorithm is capable to extract features that are more representative of the target structure, thereby enhancing the accuracy of the prediction. In the case of the fourth and fifth examples, it can be found that the incorrect predictions result from misinterpreting the structural properties of the aircraft. For example, the two-engine aircraft example in the last row of the table is predicted to be a similar category, Type J, which is a four-engine, swept-wing aircraft. This can be attributed to the reconstruction evidence where the structure of the aircraft with four engines is decoded. Hence, it is found that the predicted category is consistent to the comprehension on aircraft structure, which provides strong interpretative evidence for analyzing the model's decision.

Through the above analysis and description, we have verified the effectiveness of structure guided learning. It turns out that the structural information, including the topology configuration and key components, is of paramount importance for aircraft recognition. The proposed method enhances the recognition accuracy by facilitating a more comprehensive and accurate comprehension of the structural information.

\begin{table*}[!t]
\normalsize
\caption{\textcolor{black}{Results of different algorithms on the MT-SARD. Best results are bold.}\label{tab_3}}
\centering
\begin{tabular}{cccccccc}
\toprule
\multirow{2}{*}{Methods} & \multirow{2}{*}{\makecell{Additional \\ Annotation}} & \multicolumn{2}{c}{100\%} & \multicolumn{2}{c}{70\%} & \multicolumn{2}{c}{40\%} \\ \cmidrule(lr){3-4} \cmidrule(lr){5-6} \cmidrule(lr){7-8}
& \multicolumn{1}{c}{} & OA(\%) & Kappa & OA(\%) & Kappa & OA(\%) & Kappa \\ \midrule
SFSA \cite{RN1593} & $\times$ & 84.71 & 0.8214 & 81.51 & 0.7840 & 77.69 & 0.7401 \\ [4pt]
MGSFA-Net \cite{RN1603} & $\times$ & 82.95 & 0.8021 & 80.58 & 0.7726 & 77.17 & 0.7337 \\ [4pt]
CNN-MTL \cite{RN1590} & Angle-based & 82.95 & 0.8025 & 81.92 & 0.7901 & 78.82 & 0.7538 \\ [4pt]
STRS \cite{RN1589} & Mask-based & 85.12 & 0.8268 & 82.85 & 0.8009 & 78.31 & 0.7451 \\ [4pt]
ACM-Net \cite{RN1203} & Mask-based & 84.30 & 0.8176 & 82.13 & 0.7933 & 78.62 & 0.7515 \\ [4pt]
ESF-Net \cite{RN1043} & Mask-based & 85.43 & 0.8305 & 82.75 & 0.7994 & 79.75 & 0.7627 \\ [4pt]
\textbf{MTSGL} & Template-based & \textbf{90.21} & \textbf{0.8865} & \textbf{89.57} & \textbf{0.8791} & \textbf{85.99} & \textbf{0.8372} \\ \bottomrule
\end{tabular}
\end{table*}

\begin{figure*}[!t]
\centering
\includegraphics[width=6.5in]{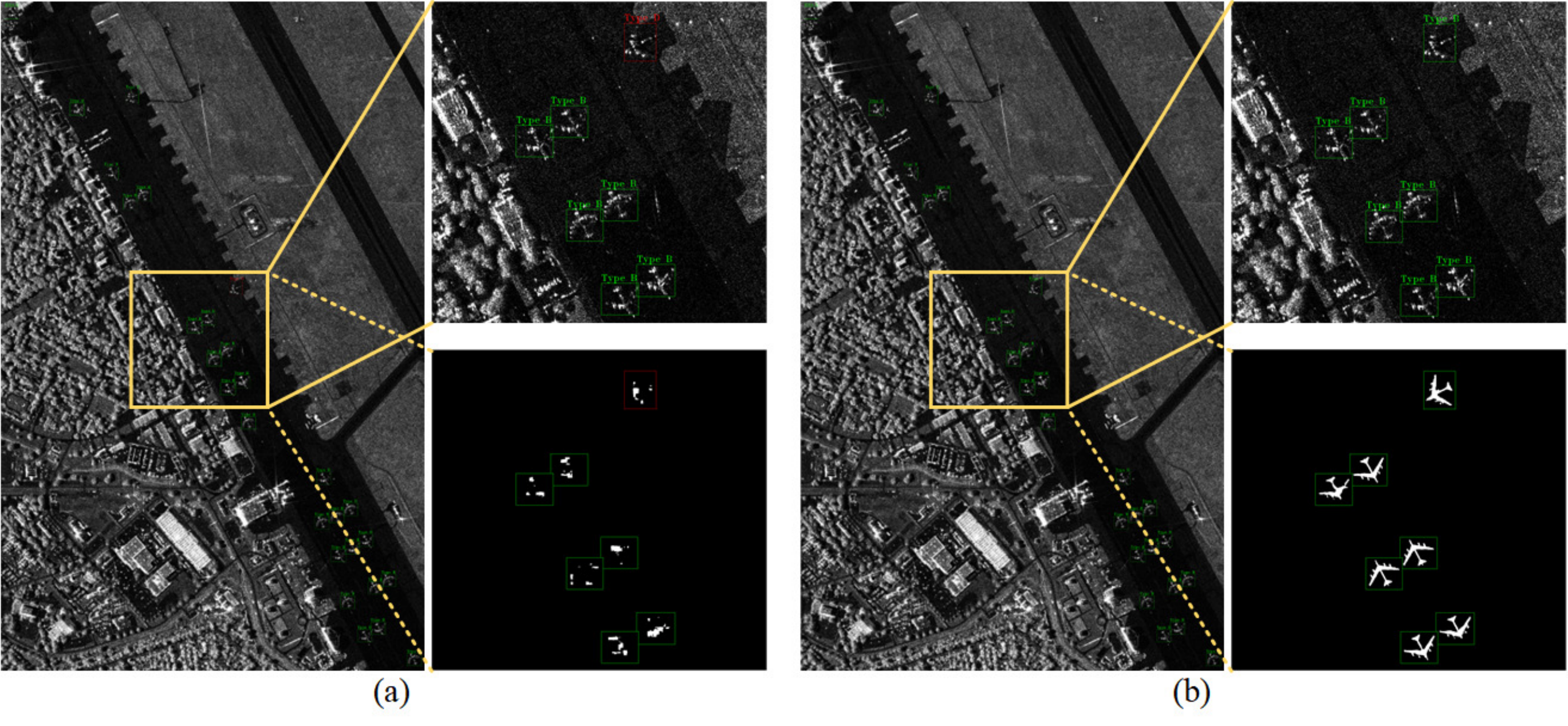}
\caption{Classification and segmentation result of the ESF-Net and the proposed method on a large scene SAR image. The correct and false classification are represented by green and red rectangles respectively. (a) ESF-Net. (b) The proposed MTSGL.}
\label{fig_15}
\end{figure*}

\subsection{Comparison With State-of-the-Art-Methods}
We choose and replicate five recent deep learning-based SAR ATR methods for comparison. Classical visual classification networks are not compared since they are not dedicated to the SAR ATR issues. Consider the SAR aircraft recognition issue, the scattering features spatial-structural association network (SFSA) \cite{RN1593} and multi-scale global scattering feature association network (MGSFA-Net) \cite{RN1603} are selected. Both methods employ a graph neural network to characterise graph structures derived from strong scattering points, which has been shown to facilitate the SAR aircraft recognition. Besides, we select the CNN with multi-task learning (CNN-MTL) \cite{RN1590}, which introduced the auxiliary task of estimating target aspect angle. Additionally, simultaneous target recognition and segmentation (STRS) \cite{RN1589}, all-convolutional based mask net (ACM-Net) \cite{RN1203} and electromagnetic scattering feature embedded network (ESF-Net) \cite{RN1043} are considered due to their adherence to the multi-task learning paradigm, where both semantic segmentation and target classification tasks are conducted together. In the original literature, the ground truths required for segmentation are obtained through manual annotation and mask extraction algorithms. To ensure a fair comparison, the masks generated by the proposed annotation method are used in their place.

The experimental results of different methods are displayed in Table \ref{tab_3}. Graph neural network-based methods perform worse than other methods because strong scattering points are inconsistent and unstable due to the discrete structure. \textcolor{black}{By comparison, the multi-task approaches based on explicit supervision have been demonstrated to enhance structural understanding, which subsequently improves the performance of classification tasks. Among them, the CNN-MTL which involves the angle-based annotation approach performs less optimally and we think this is mainly due to the orientation sparsity of our dataset. Most aircrafts are parked according to some sort of rule. The orientation of the collected aircraft images is sparsely distributed over several ranges. The sparsity property leads to degenerate solutions in training attitude estimation branch. The mask-based annotation approaches perform better because the pixel-to-pixel prediction imposes a detailed semantics comprehension on the network.} Benefiting from the structural consistency regularization and Pareto optimization, the proposed method outperforms other competitive methods and achieves the best performance in overall accuracy and Kappa coefficient under various training data size cases.

Among the comparison methods, the ESF-Net has the best performance except the proposed method due to its explicit structural guided learning. We visualize the classification result of the ESF-Net and the proposed MTSGL in a large scene SAR images, as shown in Fig. \ref{fig_15}. Besides, we utilize the mask decoder to recover the features into binary masks in order to identify the structural information embedded in the encoded features. It is evident that the vast majority of targets in this area are correctly classified in ESF-Net. The proposed method makes a better prediction, which verifies the effectiveness in practical applicability. What's more, the target structures in the decoded mask are more complete and detailed, representing more meaningful and interpretable structural features extracted by the proposed method. The structure of Type B, comprising a swept-back wing and four engines, has been faithfully reproduced. The proposed method not only provides more accurate recognition results, but also provides more reliable decision evidence. This enables interpreters and users with the opportunity to gain further insight into the accuracy of the network's understanding of the recognized aircraft.

\begin{figure}[!t]
    \centering
    \includegraphics[width=3.2in]{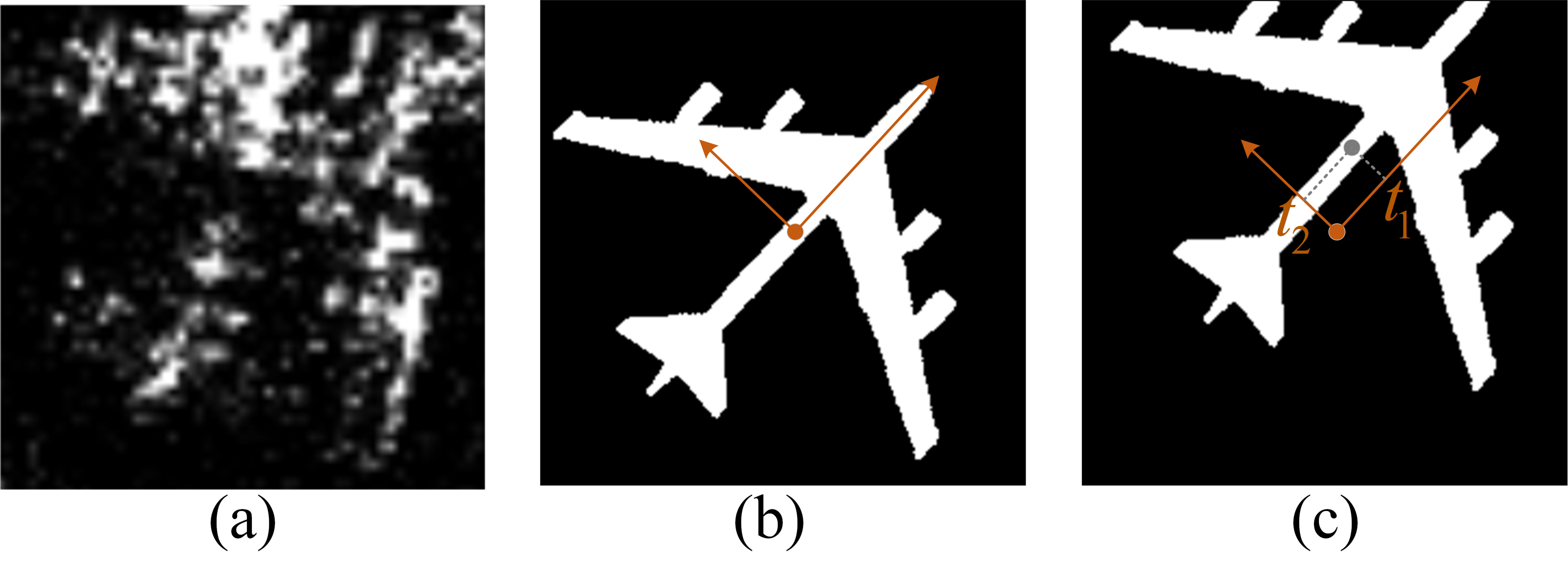}
    \caption{\textcolor{black}{The simulation of error annotation through translation on the ground truth. (a) Original image. (b) Ground truth. (c) Corrupted Annotation.}}
    \label{fig_18}
\end{figure}

\begin{table}[!t]
    \normalsize
    \caption{\textcolor{black}{Results of MTSGL trained under the corrupted datasets with various annotation error levels.}\label{tab_4}}
    \centering
    \begin{tabular}{@{} c c c @{} c @{} c @{} c}
        \toprule
        & Metric & \multicolumn{1}{c}{$\lambda=0$} & \multicolumn{1}{c}{$\lambda=0.1$} & \multicolumn{1}{c}{$\lambda=0.2$} & \multicolumn{1}{c}{$\lambda=0.3$} \\ \midrule
        \multirow{2}{*}{100\%} & OA(\%) & 90.21 & 90.01 & 88.15 & 86.5 \\
        & Kappa & 0.8865 & 0.8843 & 0.8628 & 0.8434 \\ \midrule
        \multirow{2}{*}{70\%} & OA(\%) & 89.57 & 88.4 & 86.16 & 84.99 \\
        & Kappa & 0.8791 & 0.8653 & 0.8393 & 0.8256 \\ \midrule
        \multirow{2}{*}{40\%} & OA(\%) & 85.99 & 84.4 & 81.71 & 79.96 \\
        & Kappa & 0.8372 & 0.8181 & 0.7877 & 0.7659 \\ \bottomrule
    \end{tabular}
\end{table}

\textcolor{black}{\subsection{Generalization Against the Annotation Error}}

\textcolor{black}{Note that the proposed annotation maintains a high demand of SAR aircraft expertise experience, it is essential to explore the generalization under the training dataset with annotation error. To this end, we incorporate an adjustable error factor $\lambda$ to generate labelled data at varying error levels. As shown in the Fig. \ref{fig_18}, the original ground truth (b) is corrupted into (c) by translation with ${t_1}$ in the aircraft direction and ${t_2}$ in its vertical direction. Assuming the length and width of the aircraft are ${{\rm{S}}_l}$ and ${{\rm{S}}_w}$, the translation offsets ${t_1}$ and ${t_2}$ are two random variables that are controlled by the error factor $\lambda$, sampled by
\begin{equation}
    \label{eq18}
    {t_1} \sim {\rm{uniform( - }}\lambda {{\rm{S}}_l}{\rm{,}}\lambda {{\rm{S}}_l}{\rm{),}}
\end{equation}
\begin{equation}
    \label{eq19}
    {t_2} \sim {\rm{uniform( - }}\lambda {{\rm{S}}_w}{\rm{,}}\lambda {{\rm{S}}_w}{\rm{).}}
\end{equation}}

\textcolor{black}{By controlling $\lambda$, we can simulate the corrupted training datasets with various annotation error levels. The $\lambda$ is selected at [0.1, 0.2, 0.3]. After that, the MTSGL is trained and evaluated in the manner previously outlined. The experimental results are shown in Table \ref{tab_4}. It can be seen a performance degradation with the increase of $\lambda$. Despite that, the MTSGL models trained by all levels of annotation error still perform apparently better than the baseline and the comparative methods. This inspires a certain translation tolerance that allows for robustness to the less accurate ground truth in terms of the proposed annotation.}

\section{Conclusion}
\label{section_V}
In this article, a structure-based annotation method and a mutli-task framework are designed for aircraft recognition in SAR images.
In comparison to existing annotation methods, the proposed method provides a detailed description on the aircraft structure and a global geometric transformation in accordance with the preset template. On this basis, the MTSGL is proposed, which introduces the SSA and SCR task into the target classification network to obtain expert-level understanding on aircraft in SAR images. First, the SSA task enable the network to comprehensively understand the aircraft structure. Second, the SCR task encourages learning geometrically consistent features with the preset templates' structure. Third, the Pareto-oriented optimization algorithm is introduced to yield a global multi-objective solution. We conduct extensive experiments on a self-constructed dataset MT-SARD, and the favorable outcomes substantiate the remarkable robustness and dependable interpretability of the proposed MTSGL.

\textcolor{black}{The MTSGL is model-efficient that only affects the learning process and neither changes the network architecture nor the inference process. However, it is an irrefutable fact that labelling advanced information about aircraft in SAR imagery is often expensive and labor-intensive. In recent years, the CV field has witnessed a surge of interest in weakly supervised learning for tasks such as semantic matching and object co-segmentation. Only the image-level category is needed, thereby obviating the necessity for the collection of manually labelled datasets. In can be deduced that the idea weakly supervised learning will be extended to SAR ATR in the future, which can be large beneficial for the scalability of our research work.}

\section{References Section}

\bibliographystyle{IEEEtran}

\newpage

\vspace{11pt}

\vspace{11pt}

\vfill

\end{document}